%% file: MAIN-arxiv.tex
\documentclass[letterpaper]{article} % DO NOT CHANGE THIS
\usepackage{aaai2026}  % DO NOT CHANGE THIS
\usepackage{times}  % DO NOT CHANGE THIS
\usepackage{helvet}  % DO NOT CHANGE THIS
\usepackage{courier}  % DO NOT CHANGE THIS
\usepackage[hyphens]{url}  % DO NOT CHANGE THIS
\usepackage{graphicx} % DO NOT CHANGE THIS
\usepackage{natbib}  % DO NOT CHANGE THIS AND DO NOT ADD ANY OPTIONS TO IT
\usepackage{caption} % DO NOT CHANGE THIS AND DO NOT ADD ANY OPTIONS TO IT
\usepackage{algorithm}
\usepackage{algorithmic}
\usepackage{amsthm}
\usepackage{amsmath}
\usepackage{booktabs}       % professional-quality tables
\usepackage{amsfonts}
\usepackage{float}

\usepackage{enumitem}
\usepackage{subcaption}
\usepackage{tikz}
\usetikzlibrary{positioning, arrows.meta}
\usetikzlibrary{decorations.pathreplacing, positioning}
\usepackage{multirow}
\usepackage{multicol,lipsum}

\newcommand{\wh}{\widehat}
\newcommand{\wt}{\widetilde}
\newcommand{\G}{\mathcal{G}}

\newcommand{\X}{\mathcal{X}}
\newcommand{\Y}{\mathcal{Y}}

\newcommand{\V}{\mathcal{V}}

\newcommand{\W}{\mathcal{W}}

\newcommand{\PD}{\rm{PD}}
\newcommand{\PM}{\rm{PM}}

\newcommand{\E}{\mathcal{E}}
\newcommand{\R}{\mathbb{R}}

\newcommand{\pmval}[2]{#1 {\scriptstyle \pm #2}}

\theoremstyle{plain}
\newtheorem{theorem}{Theorem}[section]

\newtheorem{lemma}[theorem]{Lemma}

\theoremstyle{definition}

\theoremstyle{remark}

\usepackage[textsize=tiny]{todonotes}
\usepackage{newfloat}
\usepackage{listings}
\DeclareCaptionStyle{ruled}{labelfont=normalfont,labelsep=colon,strut=off} % DO NOT CHANGE THIS
\floatstyle{ruled}
\newfloat{listing}{tb}{lst}{}
\floatname{listing}{Listing}
 \usepackage[capitalize,noabbrev]{cleveref}

\title{T3former: Temporal Graph Classification\\ with Topological Machine Learning}

\author {
    Md.~Joshem~Uddin\textsuperscript{\rm 1},
    Soham~Changani\textsuperscript{\rm 1},
    Baris~Coskunuzer\textsuperscript{\rm 1}
}
\affiliations {
    \textsuperscript{\rm 1}Department of Mathematical Sciences, The University of Texas at Dallas,     Richardson, TX, USA\\
    MdJoshem.Uddin@utdallas.edu, soham.changani@utdallas.edu, coskunuz@utdallas.edu
}

\usepackage{bibentry}
\begin{document}

\maketitle

\begin{abstract}
Temporal graph classification plays a critical role in applications such as cybersecurity, brain connectivity analysis, social dynamics, and traffic monitoring. Despite its significance, this problem remains underexplored compared to temporal link prediction or node forecasting. Existing methods often rely on snapshot-based or recurrent architectures that either lose fine-grained temporal information or struggle with long-range dependencies. Moreover, local message-passing approaches suffer from oversmoothing and oversquashing, limiting their ability to capture complex temporal structures.

We introduce \textsc{T3former}, a novel Topological Temporal Transformer that leverages sliding-window topological and spectral descriptors as first-class tokens, integrated via a specialized Descriptor-Attention mechanism. This design preserves temporal fidelity, enhances robustness, and enables principled cross-modal fusion without rigid discretization.
T3former achieves state-of-the-art performance across multiple benchmarks, including dynamic social networks, brain functional connectivity datasets, and traffic networks. It also offers theoretical guarantees of stability under temporal and structural perturbations. Our results highlight the power of combining topological and spectral insights for advancing the frontier of temporal graph learning.

\end{abstract}

\vspace{-.1in}
\begin{links}
    {\footnotesize \link{Code}{https://anonymous.4open.science/r/T3Former-3311}}
    %\link{Datasets}{https://aaai.org/example/datasets}
    %\link{Extended version}{https://aaai.org/example/extended-version}
\end{links}

\vspace{-.1in}
\section{Introduction} \label{sec:intro}

\input{sections/01-intro}

\section{Background} \label{sec:background}
\input{sections/02-background}

\section{Methodology} \label{sec:methodology}

\input{sections/03-methodology}

\section{Experiments} \label{sec:experiments}

\input{sections/04-experiments}

\section{Conclusion} \label{sec:conclusion}
We introduced \textbf{T3former}, a temporal graph classification framework that unifies global topological invariants, spectral descriptors, and local structural features via a transformer-based \emph{Descriptor-Attention} mechanism. Extensive experiments across diverse temporal domains     highlight both the adaptability of T3former and the complementary strengths of its descriptors, demonstrating superior performance. Promising future directions include extending T3former to streaming and continuous-time settings without relying on predefined sliding windows, exploring adaptive temporal resolutions, and incorporating additional topological invariants for richer feature extraction.

% We introduced \textbf{T3former}, a temporal graph classification framework that integrates global topological invariants, spectral descriptors, and local structural features within a unified transformer-based \emph{Descriptor-Attention} mechanism. By adaptively fusing these complementary modalities, T3former effectively captures both fine-grained temporal patterns and higher-order graph structure. Extensive experiments across diverse domains—including social, traffic, and brain networks—demonstrate that T3former consistently outperforms existing baselines and delivers state-of-the-art results.

% Looking ahead, promising research directions include extending T3former to handle streaming and continuous-time scenarios without predefined sliding windows, exploring adaptive temporal resolutions for greater flexibility, and incorporating additional topological invariants for richer graph representations. We believe these advances will further broaden the applicability of T3former and stimulate future research in temporal graph learning.

\clearpage
\bibliography{references}

\clearpage

\appendix

\centerline{\bf \LARGE Appendix}
%\onecolumn
\input{sections/05-appendix}

\end{document}

%% file: sections/01-intro.tex
Temporal graph classification, the task of assigning a label to a single temporal graph whose nodes and edges carry timestamp information indicating their active intervals, plays a crucial role in domains such as cybersecurity intrusion detection, dynamic functional‑connectivity mapping in neuroscience, social network analysis, and traffic pattern recognition. Current approaches predominantly rely on discrete snapshot methods or recurrent updates (e.g., \textsc{EvolveGCN} \cite{trivedi2019dyrep}, \textsc{DySAT} \cite{zhao2020dysat}), inherently trading off between temporal granularity and the global structural context. Local message-passing networks further suffer from critical limitations like \emph{oversmoothing} and \emph{oversquashing}, restricting their capacity to identify long-range and higher-order temporal patterns \cite{alon2021bottleneck}. Moreover, the learned embeddings of these methods are frequently unstable under small perturbations, a notable disadvantage in inherently noisy and sparse settings such as neuroscience \cite{hajij2021expressive}.

In contrast, static graph classification techniques have recently made significant progress by leveraging global topological and spectral descriptors, such as persistent homology \cite{immonen2024going,hiraoka2024topological} and Laplacian density-of-states (DOS) curves \cite{dong2019network}, providing robust and expressive graph-level summaries. However, extending these methods directly to temporal settings typically demands rigid snapshotting, compromising temporal resolution.

To bridge this gap, we introduce \textsc{T3former}, a novel \emph{Topological Temporal Transformer} specifically designed for temporal graph classification. \textsc{T3former} effectively integrates topological signatures and Laplacian DOS vectors through sliding-windows as first-class descriptor tokens alongside global structural tokens generated by lightweight per-window graph neural networks. 
%\JU{can we add this in abstract: as first class token alongside....}. 
These tokens, enriched with relative-time embeddings, are effectively fused through a unified transformer backbone with dedicated \emph{Descriptor-Attention} modules, explicitly capturing intricate interactions among structural, topological, and spectral modalities.

The \emph{Descriptor-Attention} framework offers significant advantages: (1) it maintains fine-grained temporal resolution without ad-hoc snapshot discretization; (2) it enhances robustness through topological invariants; (3) it leverages mesoscopic structural insights provided by spectral descriptors; and (4) it ensures principled cross-modal integration via transformer-based attention rather than simple concatenation.
Empirically, we demonstrate the superior performance of \textsc{T3former} across standard benchmarks spanning social networks, dynamic brain connectivity graphs, and traffic networks, establishing new state-of-the-art results. Our extensive experiments further confirm the method's robustness, interpretability, and resilience to perturbations.

In summary, our contributions include:
\begin{itemize}
\item \textsc{T3former}, a novel temporal graph classification model that integrates sliding-window topological and spectral descriptors via dedicated cross-attention modules.
\item \textit{Descriptor-Attention}, a transformer-based fusion mechanism that unifies structural, topological, and spectral information.
%\item A sliding-window framework that preserves fine-grained temporal dynamics while leveraging powerful graph-level summaries.
\item Theoretical stability guarantees showing bounded sensitivity of topological and spectral descriptors to minor graph perturbations.
\item We introduce new temporal graph classification datasets adapted from real-world traffic networks.
%\SC{I added this here. I think it's part of our contributions too.}
\item Extensive experiments on social, brain connectivity, and traffic network benchmarks, demonstrating state-of-the-art performance, robustness, and broad applicability.
\end{itemize}

%% file: sections/02-background.tex
\subsection{Related Work}

\paragraph{Temporal Graph Neural Networks.}
Temporal graph neural networks (TGNNs) and graph transformers have emerged as a popular framework for modeling dynamic graphs, where nodes, edges, and features evolve over time. A common strategy in TGNNs is to partition the temporal graph into discrete snapshots, apply graph neural networks (GNNs) to obtain node or graph representations for each snapshot, and then model the temporal evolution of these representations using sequence models such as recurrent neural networks (RNNs) or transformers. Representative examples include \textsc{EvolveGCN}~\cite{pareja2020evolvegcn}, which adapts GCN parameters through recurrent updates, and \textsc{Dygformer}~\cite{yu2023towards}, which models dynamic events through temporal point processes. More recent methods such as \textsc{TGN}~\cite{rossi2020temporal} leverage memory modules to capture fine-grained temporal dependencies between node interactions. While effective for dynamic prediction tasks such as link prediction or event forecasting, these methods often rely heavily on local neighborhood aggregation and sequential modeling of embeddings, making them sensitive to temporal noise and structural perturbations.

\paragraph{Topological Methods for  Temporal Graph Learning.}
An alternative to sequence modeling over learned embeddings is to focus on improving the snapshot representations themselves using topological methods. Topological models utilizing persistent homology have recently become a strong alternative to GNNs, especially in graph-level learning~\cite{loiseaux2024stable,chen2024topogcl,verma2024topological} as they have proven highly effective for capturing higher-order structural patterns in static graphs.  In the context of dynamic or temporal graphs, recent works such as TAMP-\textsc{S2GCNets}~\cite{chen2022tamp}, \textsc{GraphPulse}~\cite{shamsi2024graphpulse}, and Dynamic Dowker Filtrations~\cite{ye2023stable} demonstrate that topological summaries can serve as powerful snapshot encoders, significantly boosting performance and often outperforming traditional TGNN architectures. Topological methods capture long-range dependencies and temporal dynamics by modeling global and higher-order patterns, without relying solely on sequential updates.

\paragraph{Prediction-Focused Temporal Learning.}
Despite these advances, most temporal learning methods remain focused on prediction tasks such as dynamic link prediction, node classification, or next-event forecasting. Models like \textsc{Jodie}~\cite{kumar2019predicting}, \textsc{DyRep}~\cite{trivedi2019dyrep}, \textsc{TGN}~\cite{rossi2020temporal}, and \textsc{APAN}~\cite{wang2021apan} primarily target node- or edge-level predictions over time. Recent benchmarking efforts such as the Temporal Graph Benchmark (TGB)~\cite{huang2023temporal} also emphasize predictive tasks across diverse domains, highlighting the relative scarcity of methods designed specifically for graph-level classification in temporal settings. Furthermore, sequence-based models often exhibit sensitivity to timestamp irregularities and noise, which can degrade the learned representations when applied to tasks requiring robust global graph understanding.

% \paragraph{Temporal Graph Classification.}
% While most temporal graph learning models have been studied for temporal graph property prediction tasks, temporal graph classification remains comparatively underexplored~\cite{kim2022graph,cai2021structural}. Only a few recent works have adapted TGNNs for graph-level tasks. Most existing methods cannot effectively utilize temporal information in the learning process, as they struggle when fine-grained temporal evolution cannot be easily discretized~\cite{ekle2024anomaly}. Recently, \cite{tieu2024temporal} brought attention to this important problem by introducing several benchmark datasets, and baseline methods.  Our work addresses this gap by introducing a hybrid framework that captures both global structural patterns and continuous temporal evolution using topological and spectral summaries, enabling effective classification of evolving graphs without rigid snapshot assumptions.

\paragraph{Temporal Graph Classification.}
While much of the research on temporal graph learning has centered around \textit{temporal graph property prediction }tasks, the problem of \textit{temporal graph classification} remains relatively underexplored~\cite{kim2022graph}. Only a few recent efforts have extended TGNNs to temporal graph classification~\cite{yang2025generalizable,cai2021structural}, and many existing approaches struggle to effectively leverage fine-grained temporal information, particularly when the dynamic evolution cannot be easily discretized into snapshots~\cite{ekle2024anomaly}. Recognizing this gap, \cite{tieu2024temporal} recently introduced a suite of benchmark datasets and baseline methods to promote further study of temporal graph classification. Nevertheless, most current models continue to rely heavily on rigid discretization schemes or sequence-based assumptions, limiting their ability to capture continuous structural evolution. Our work addresses these limitations by proposing a hybrid framework that integrates global structural modeling with timestamp-aware topological and spectral summaries, enabling effective classification of evolving graphs without requiring fixed snapshot sequences.

\subsection{Temporal Graph Learning}
Many real-world systems are inherently dynamic, where the graph structure evolves over time through the addition or removal of nodes and edges. To capture such temporal behavior, graphs can be extended to the temporal domain using either continuous-time or discrete-time representations. 

Let a static undirected graph be defined as $\G = (\V, \E)$, where $\V$ is the set of nodes and $\E$ is the set of edges. Each node $v \in \V$ is associated with a feature vector $\mathbf{x}_v \in \mathbb{R}^d$, and each edge $(u, v) \in \E$ has a corresponding feature $\mathbf{e}_{uv}$. The neighborhood of a node $v$ is denoted by $\mathcal{N}(v)$.

% To model evolving graphs, a \textit{continuous-time dynamic graph} (CTDG) represents the graph as a sequence of timestamped interactions: 
A \textit{continuous-time dynamic graph} (CTDG) models evolving graphs as a sequence of timestamped interactions:

\noindent $\G = \{(u, v, \mathbf{x}_u(t), \mathbf{x}_v(t), \mathbf{e}_{uv}(t), t)\}_{t = t_0},$
where each event indicates that nodes $u$ and $v$ interacted at time $t$, and $\mathbf{x}_u(t)$, $\mathbf{x}_v(t)$, and $\mathbf{e}_{uv}(t)$ are the corresponding node and edge features at that time.

% For computational tractability, the CTDG can be discretized into a \textit{discrete-time dynamic graph} (DTDG), represented as a sequence of graph snapshots $\{\G^{(t)}\}$. A snapshot at time $t \geq t_0$ is denoted as:
For computational tractability, a CTDG can be discretized into a \textit{discrete-time dynamic graph} (DTDG), represented as snapshots ${\G^{(t)}}$, where each snapshot at time $t \geq t_0$ is:

\noindent $\G^{(t)} = \{(u, v, \mathbf{x}_u(\bar{t}), \mathbf{x}_v(\bar{t}), \mathbf{e}_{uv}(\bar{t}), \bar{t})\}_{\bar{t} = t_0}^{t},$
which aggregates all events up to time $t$ by updating the initial graph $\G^{(t_0)}$ using the event stream. 
The temporal graph is defined by the last snapshot, $\G = \G^{(T)}$, where $T$ is the final timestamp. Let $\V^{(t)}$ and $\E^{(t)}$ be the node and edge sets at time $t$. The temporal neighborhood of node $v$ at time $t$ is:
% The final temporal graph corresponds to the last snapshot, i.e., $\G = \G^{(T)}$ where $T$ is the maximum timestamp.
% Let $\V^{(t)}$ and $\E^{(t)}$ denote the node and edge sets at time $t$. The temporal neighborhood of a node $v$ at time $t$ is defined as:

\noindent $\mathcal{N}^{(t)}(v) = \{(u, t): (u, v, \mathbf{x}_u(\bar{t}), \mathbf{x}_v(\bar{t}), \mathbf{e}_{uv}(\bar{t})) \in \G^{(t)}\},$
including all nodes $u$ that interacted with $v$ at any time $\bar{t} \leq t$.
%Throughout this work, we use $\G$ to refer to the entire temporal graph and $\G^{(t)}$ to denote a snapshot at time $t$.

\subsection{Persistent Homology}
\label{sec:PH}

Persistent Homology (PH) is a central tool in Topological Data Analysis (TDA) that captures multiscale topological features such as connected components, loops, and voids that persist across different scales~\cite{dey2022computational}. Originally developed for point cloud data, PH has since been successfully extended to graphs, images, and other data modalities~\cite{coskunuzer2024topological}. In graph learning, PH provides latent structural insights that complement conventional methods and GNNs, enabling a richer understanding of both local and global graph structure.

\begin{figure}[h!]
\centering
\includegraphics[width=.8\linewidth]{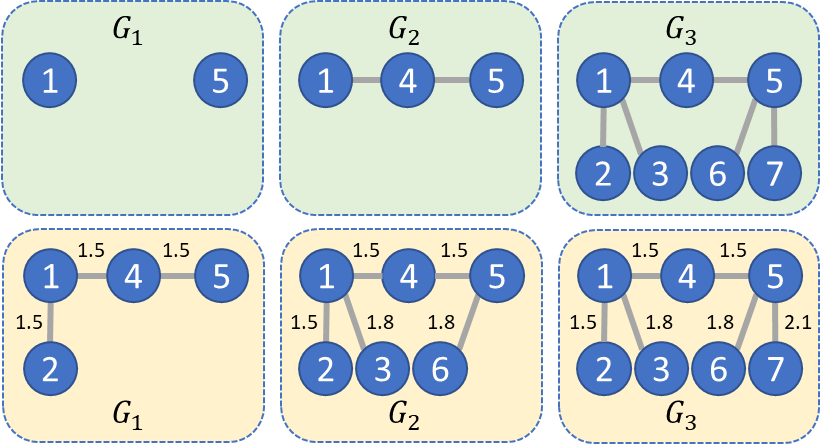}
\caption{\footnotesize \textbf{Graph Filtration.} For $\mathcal{G}=\mathcal{G}_3$ in both examples, the top figure illustrates a \textit{superlevel filtration using the node degree function} with thresholds $3>2>1$, where nodes of degree 3 are activated first, followed by those of lower degrees.  Similarly, the bottom figure illustrates a \textit{sublevel filtration based on edge weights} with thresholds $1.5< 1.8< 2.1$.  \label{fig:graph-filtration}}
\end{figure}

PH typically consists of three main steps: constructing a filtration, obtaining persistence diagrams, and performing vectorization. The process begins with \textit{filtration}, where a nested sequence of simplicial complexes is built to track the evolution of topological features. Given an unweighted graph $\mathcal{G} = (\mathcal{V}, \mathcal{E})$ and a filtration function $f: \mathcal{V} \to \mathbb{R}$ with thresholds $\mathcal{I} = \{\alpha_i\}$, we construct subgraphs $\mathcal{G}^i = (\mathcal{V}_i, \mathcal{E}_i)$, where $\mathcal{V}_i = \{v \in \mathcal{V} \mid f(v) \leq \alpha_i\}$ (See~\Cref{fig:graph-filtration} for toy example). The filtration function can be based on graph structural properties (e.g., degree, betweenness centrality) or derived from domain-specific information (e.g., atomic number in molecular graphs). For weighted graphs, edge weights can also serve as a natural filtration~\cite{coskunuzer2024topological}. The primary goal is to establish a hierarchy among nodes or edges that yields a meaningful sequence of subgraphs aligned with the downstream task. In temporal graphs, a natural choice would be timestamps on edges as a valuable filtration function.

Once the subgraphs are defined, each $\mathcal{G}^i$ is extended to its clique complex $\widehat{\mathcal{G}}^i$, producing a sequence $\widehat{\mathcal{G}}^1 \subseteq \widehat{\mathcal{G}}^2 \subseteq \dots \subseteq \widehat{\mathcal{G}}^N$. As the filtration evolves, topological features such as connected components (0-holes), loops (1-holes), and cavities (2-holes) emerge and disappear. Each $k$-dimensional feature $\sigma$ is associated with a birth index $b_\sigma$ and death index $d_\sigma$ ($1 \leq b_\sigma < d_\sigma \leq N$), recorded as a point $(b_\sigma, d_\sigma)$ in the $k$-th persistence diagram ${\rm PD}_k(\mathcal{G}) = \{(b_\sigma, d_\sigma) \mid \sigma \in H_k(\widehat{\mathcal{G}}^i)\}$. Since persistence diagrams are sets of points in $\mathbb{R}^2$, a \textit{vectorization} step is necessary for compatibility with machine learning models. Techniques such as persistence images, landscapes, silhouettes, and Betti curves~\cite{ali2023survey} are commonly used to transform diagrams into fixed-size feature representations.

While PH effectively captures complex structural patterns, its integration into temporal graph learning remains challenging. Most prior methods are limited to static graphs and fail to model evolving topological features. Additionally, combining PH summaries with temporal dynamics and node attributes is nontrivial. To address these issues, we propose a hybrid framework that extracts topological descriptors from temporal subgraphs, aligning them with sliding windows to track structural changes over time. This enables scalable, interpretable temporal graph classification without relying on rigid snapshots or heavy discretization.

%In this work, we extend traditional single-parameter vectorizations by emphasizing Betti numbers $\beta_i$, which count the number of $i$-dimensional topological features at each scale. By applying PH across temporal subgraphs and tracking Betti curves over sliding time windows, we capture both structural and temporal evolution, enabling finer-grained analysis of evolving graphs. This approach naturally aligns with temporal graph classification tasks, providing robust and interpretable topological summaries without relying on rigid snapshot discretization. 

% Despite its strength in capturing complex structural patterns, integrating PH into temporal graph learning remains challenging. Most existing methods focus on static graphs and do not naturally extend to evolving structures where topological features change over time. Moreover, traditional approaches often struggle to combine topological summaries with temporal dynamics and node attribute evolution. Motivated by these challenges, our work proposes a new hybrid framework that incorporates topological descriptors derived from PH over temporal subgraphs, providing a scalable and interpretable way to track structural changes over time. By aligning persistent homology features with temporal windows, our approach enables effective temporal graph classification without relying on rigid snapshot sequences or heavy discretization assumptions.

\subsection{Density of States for Graphs}

The \textit{density of states} (DoS) is a classical concept from spectral graph theory that captures the distribution of eigenvalues of a graph Laplacian~\cite{chung1997spectral}. It provides a coarse-grained summary of the graph’s structural complexity, bottlenecks, and connectivity properties.

Given an undirected graph $G = (\mathcal{V}, \mathcal{E})$, let $L$ denote its normalized Laplacian matrix. The eigenvalues of $L$, denoted by $\{\lambda_i\}_{i=1}^{|\mathcal{V}|}$, lie within the interval $[0,2]$. The DoS is then defined as the empirical distribution of these eigenvalues: \quad \quad \quad
\centerline{$\displaystyle{\text{DoS}(\lambda) = \dfrac{1}{|\mathcal{V}|} \sum_{i=1}^{|\mathcal{V}|} \delta(\lambda - \lambda_i)}$}
where $\delta(\cdot)$ denotes the Dirac delta function.

In practice, the DoS is approximated by constructing a histogram over a fixed number of bins partitioning $[0,2]$~\cite{dong2019network}. This yields a compact and interpretable vector representation that captures key spectral characteristics of the graph. The resulting descriptor is permutation-invariant, robust to small perturbations in the graph structure, and encodes information about fundamental graph properties such as connectivity, expansion, and community structure~\cite{spielman2011spectral}.
When applied to temporal graphs, the DoS computed over sliding subgraphs provides a dynamic signature of the evolving spectral complexity~\cite{huang2024laplacian}, complementing topological features.

%% file: sections/03-methodology.tex
We propose a hybrid framework that integrates global structural modeling with timestamp-aware topological and spectral descriptors for temporal graph classification. Our approach is composed of four key components: temporal filtration via sliding windows, topological and spectral descriptor extraction, static graph construction with temporal node features, and a unified classification architecture that combines multiple embeddings through self-attention.

\subsection{Temporal Filtration via Sliding Window}

A fundamental obstacle in temporal graph classification is that interactions arrive as continuously timestamped events rather than as a predetermined sequence of discrete snapshots. To address this, we introduce a \emph{temporal filtration} framework inspired by persistent homology, in which the timestamp annotation itself induces a filtration over the underlying static graph.

Formally, let $\mathcal{G} = (\mathcal{V}, \mathcal{E}, \tau)$
denote a temporal graph with node set $\mathcal{V}$, edge set $\mathcal{E}$, and the relation $\tau\!:\mathcal{E}\to\mathbb{R}^+$ assigning each edge its timestamp (or multiset of timestamps for repeated interactions). Given a window length $\delta>0$, we define a family of subgraphs $G_{[t,\,t+\delta]} =\bigl(\mathcal{V}_t,\mathcal{E}_t\bigr)$, 
where $\mathcal{E}_t = \{e\in\mathcal{E}\mid \tau(e)\in[t,t+\delta]\}$ and  $\mathcal{V}_t = \{\,u\in\mathcal{V}\mid \exists\,e\in\mathcal{E}_t \mbox{ with }  u\in e\}$ (See~\Cref{fig:temporal} for toy example). To control the granularity of the sliding windows, we also use a stride parameter $\sigma \in (0, \delta)$ that defines the step size of the windowing process, resulting in the filtered sequence $\G_i=\G_{[i\sigma,\,i\sigma+\delta]}$ (See~\Cref{app:sliding-window}). 

This sliding-window construction yields a pseudo-filtration in time, where each edge may naturally appear in multiple windows if it carries multiple timestamps. Crucially, our approach avoids collapsing an edge to a single timestamp (e.g.\ $\min$, $\max$, or average), which can obscure temporal patterns. By avoiding coarse snapshotting, temporal filtration preserves high-resolution dynamics and yields a structured subgraph sequence that captures both local transitions and global evolution.

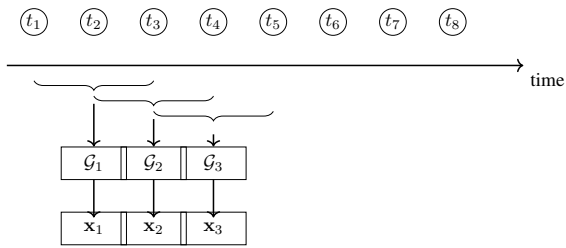
\begin{figure}[h!]
%\vspace{-.2in}
\resizebox{.9\linewidth}{!}{
\begin{tikzpicture}[scale=1, every node/.style={scale=1}]

    % Time series points with circles
    \foreach \x/\label in {0/$t_1$, 1/$t_2$, 2/$t_3$, 3/$t_4$, 4/$t_5$, 5/$t_6$, 6/$t_7$, 7/$t_8$} {
        \node[draw, circle, inner sep=1pt] (n\x) at (\x*1.1,0) {\label};
    }
    
    % Time axis
    \draw[->, thick] (-0.5,-0.8) -- (9,-0.8) node[below right] {time};

    % Sliding windows braces (slightly more space)
    \draw[decorate,decoration={brace,mirror,amplitude=5pt},yshift=-8pt]
        (0,-0.8) -- (2.2,-0.8) node[midway, yshift=-0.3cm] (b1) {}; % t1-t3
    \draw[decorate,decoration={brace,mirror,amplitude=5pt},yshift=-16pt]
        (1.1,-0.8) -- (3.3,-0.8) node[midway, yshift=-0.3cm] (b2) {}; % t2-t4
    \draw[decorate,decoration={brace,mirror,amplitude=5pt},yshift=-24pt]
        (2.2,-0.8) -- (4.4,-0.8) node[midway, yshift=-0.3cm] (b3) {}; % t3-t5

    % Graphs G1, G2, G3 aligned horizontally (slightly lower)
    \node[draw, rectangle, minimum width=12mm, minimum height=6mm] (g1) at (1.1,-2.6) {$\G_1$};
    \node[draw, rectangle, minimum width=12mm, minimum height=6mm] (g2) at (2.2,-2.6) {$\G_2$};
    \node[draw, rectangle, minimum width=12mm, minimum height=6mm] (g3) at (3.3,-2.6) {$\G_3$};

    % Straight arrows from brace centers to graphs
    \draw[->, thick] (b1) -- (g1.north);
    \draw[->, thick] (b2) -- (g2.north);
    \draw[->, thick] (b3) -- (g3.north);

    % Arrows from G to x
    \foreach \x in {1.1,2.2,3.3} {
        \draw[->, thick] (\x,-2.9) -- (\x,-3.6);
    }

    % Feature vectors x1, x2, x3 aligned horizontally
    \node[draw, rectangle, minimum width=12mm, minimum height=6mm] at (1.1,-3.8) {$\mathbf{x}_1$};
    \node[draw, rectangle, minimum width=12mm, minimum height=6mm] at (2.2,-3.8) {$\mathbf{x}_2$};
    \node[draw, rectangle, minimum width=12mm, minimum height=6mm] at (3.3,-3.8) {$\mathbf{x}_3$};

\end{tikzpicture}}
\caption{\footnotesize \textbf{Sliding window construction.} DoS and Betti vectors $\{\X_i\}$) are extracted for each induced subgraph $\G_t=\G_{[t,t+\delta]}$.}
%\vspace{-.2in}
\end{figure}

%\BC{Soham, add a figure illustrating the sliding window construction.}

% Formally, given a temporal graph $\mathcal{G} = (\mathcal{V}, \mathcal{E}, \tau)$, where $\mathcal{V}$ is the set of nodes, $\mathcal{E}$ the set of edges, and $\tau: \mathcal{E} \to \mathbb{R}^+$ assigns timestamps to edges, we define a sliding window of size $\delta > 0$. For each time interval $[t, t+\delta]$, we extract a subgraph:
% \[
% G_{[t, t+\delta]} = (\mathcal{V}_t, \mathcal{E}_t), \quad \text{where} \quad \mathcal{E}_t = \{ e \in \mathcal{E} \mid \tau(e) \in [t, t+\delta] \},
% \]
% and $\mathcal{V}_t$ includes all nodes incident to edges in $\mathcal{E}_t$.

\subsection{T3Former}

We present \textsc{T3Former}, a \textit{Topological Temporal Transformer} framework that effectively incorporates topological and spectral signatures of the temporal graph into a unified graph transformer architecture.

\paragraph{Topological Descriptor Vectors.}
For each subgraph $\G_{[t, t+\delta]}$, we compute a \textit{topological descriptor vector}:

\noindent $\phi_t = \left[ |\mathcal{V}_t|, |\mathcal{E}_t|, \beta_0(\wh{G}_{[t, t+\delta]}), \beta_1(\wh{G}_{[t, t+\delta]}) \right],$ \quad 
where $\wh{G}_{[t, t+\delta]}$ denotes the clique complex of $\G_{[t, t+\delta]}$, and $\beta_0$, $\beta_1$ denote the zeroth and first Betti numbers, respectively. These features capture key aspects of connectivity and cyclic structure over time, providing interpretable and noise-robust summaries of the evolving graph topology.

\paragraph{Spectral Descriptor Vectors.}
In parallel, we capture spectral properties of each subgraph using its \textit{density of states} (DOS) vector. Given the normalized Laplacian $L_t$ of $\G_{[t, t+\delta]}$, the DOS is approximated by a histogram over the eigenvalues: \quad $\psi_t = \text{Histogram}(\text{Eigenvalues}(L_t)).$

The DOS provides a complementary, geometry-sensitive representation of the graph, encoding notions such as graph complexity, bottlenecks, and expansion properties.

\paragraph{Global Structural Modeling via Graph Neural Network.}
While the topological and spectral descriptors extracted through the sliding window model fine-grained temporal dynamics, we also capture global structural patterns by applying a Graph neural network GSAGE~\cite{hamilton2017inductive} to the full static graph $\mathcal{G}$ (ignoring timestamps). To incorporate temporal activity, we assign each node a binary feature vector of length $|\tau|$, \textit{temporal degree} (see~\Cref{app:datasets}), where each entry in a node’s feature vector is set to 1 if the node is involved in at least one edge at that time step, and 0 otherwise. This encoding enables the static graph to reflect temporal behavior while preserving its overall structure. Given initial node features (e.g., temporal degree, clustering coefficient, or learned embeddings), GSAGE generates node embeddings, aggregated via global pooling into a holistic graph-level representation. 

%\SC{Should we make it clear here that we only use temporal degree for social networks since there were no node features?} \BC{We cannot talk about datasets before experiments. We mention in the hyperparameters, etc. part}
%\JU{table 1 (feature)+ both is mentioned on model configuration }

% \paragraph{Integration and Classification.}
% Both the topological ($\phi_t$) and spectral ($\psi_t$) descriptors  are fed into a Transformer encoder, producing two additional streams of temporal embeddings that capture localized structural dynamics. Alongside these, the temporal embeddings from the GSAGE provide a global structural perspective. To unify these three complementary representations, we apply a self-attention mechanism that adaptively weighs and fuses the embeddings. The resulting fused embedding is passed through a linear classification layer to predict the label associated with the input temporal graph.

\paragraph{Integration and Classification.}
Topological ($\phi_t$) and spectral ($\psi_t$) descriptors are processed through a Transformer encoder, yielding two streams of temporal embeddings that capture localized structural dynamics. These are combined with global embeddings from GSAGE. A self-attention mechanism adaptively fuses the three views, and the resulting embedding is passed to a linear layer for final graph classification.

\begin{figure}[t]
\centering
\includegraphics[width=\linewidth]{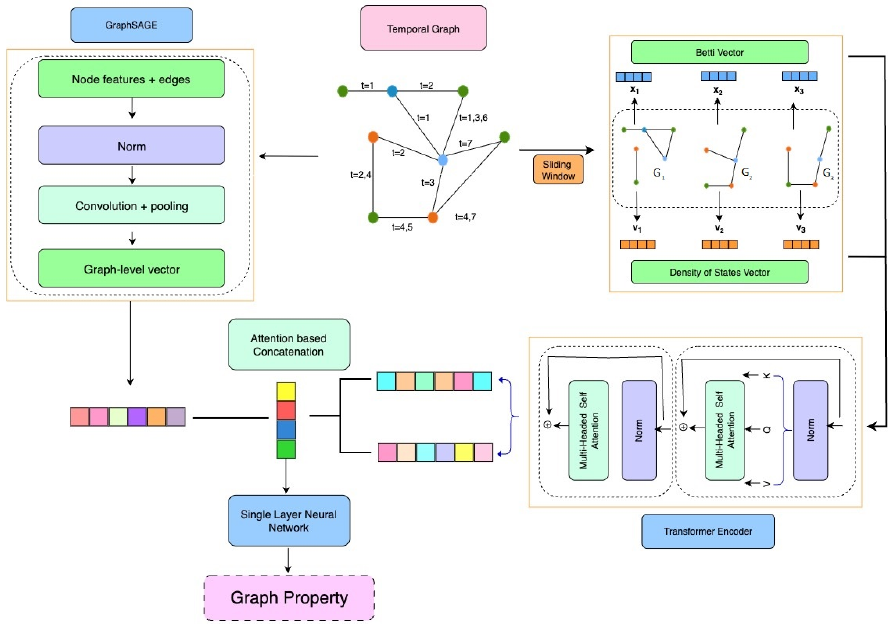}
\caption{\footnotesize \textbf{T3Former Flowchart.}  Given a temporal graph $\G=(\V,\E,\tau)$, GraphSAGE generates time-aware structural embeddings. In parallel, a sliding-window module extracts topological (Betti) and spectral (density of states) vectors. These sequential features are processed through a Transformer to capture temporal dependencies. An attention-based fusion module merges the multi-view representations, which are then passed through a final neural layer for graph property prediction.}
\label{fig:t3former}
%\vspace{-.2in}
\end{figure}

The \textsc{T3former} architecture (\Cref{fig:t3former}) uses GraphSAGE to encode time‑agnostic global structure, applies a sliding‑window filtration to extract timestamp‑aware topological and spectral features, and fuses these views via attention and a transformer encoder, capturing both high‑level global structures and fine‑grained temporal dynamics in a scalable framework for temporal graph classification.

%The \textsc{T3former} architecture (\Cref{fig:t3former}) integrates a global graph transformer encoder with timestamp-aware topological and spectral descriptors derived from temporal filtrations. This unified design enables \textsc{T3former} to jointly capture high-level structural patterns and fine-grained temporal dynamics, providing a scalable and principled solution for temporal graph classification.

% This unified architecture enables \textsc{T3former} to simultaneously capture both high-level global structures and detailed temporal dynamics, offering a scalable and principled solution for temporal graph classification.

% The \textsc{T3former} architecture is shown in~\Cref{fig:t3former}. The model integrates a global graph transformer encoder with timestamp-aware topological and spectral descriptors extracted through temporal filtration, allowing it to simultaneously capture structural and temporal dynamics of evolving graphs.

\subsection{Stability of Topological and Spectral Descriptors} \label{sec:stability}

For temporal graph classification, robustness to small perturbations in the graph structure is critical, particularly in real-world settings where data may be noisy, incomplete, or prone to timestamp inaccuracies. In \textsc{T3former}, we utilize topological and spectral descriptors that are inherently stable: small changes to the graph, such as edge insertions or deletions for given time periods induce only bounded changes in the extracted features. This stability ensures that the learned representations are resilient to noise and minor fluctuations over time, which is essential for reliable temporal graph classification. 

We formalize these stability properties below. First, we prove the stability of our topological descriptors. Notice that our theorem is more general, and it applies to any sliding window $\G_{[t,t+\delta]}$.

\begin{theorem}[Stability of Topological Descriptors] \label{thm:stability-topo} Let $\G=(\V,\E) $ be a graph and let $\tau_1,\tau_2 :\E\to\R$ be two timestamp functions on $\G$. Then, for $k\geq 0$, we have 
$$\|\vec{\beta}_k(\G,\tau_1)-\vec{\beta}_k(\G,\tau_2)\|_1 \leq C_k\cdot \|\tau_1-\tau_2\|_1$$
where $\vec{\beta}_k(\G,\tau_i)$ represents the Betti vector corresponding to $\PD_k(\G,\tau_i)$ obtained by sublevel filtration with respect to $\tau_i$.
\end{theorem}

Next, we show the stability of our spectral descriptors.

\begin{theorem}[Stability of Spectral Descriptors] \label{thm:stability-dos}
Let $\mathcal{G}$ and $\mathcal{G}'$ be two temporal graphs differing by at most $k$ edge modifications (insertions or deletions) within any window $[t, t+\delta]$. Then, the Wasserstein distance between the spectral descriptors $\psi_t$ and $\psi'_t$ is bounded by $Ck/n$, where $n = |\mathcal{V}|$ and constant $C$ depends on the eigenvalue distribution.
\end{theorem}

The proofs of the theorems are given in~\Cref{sec:proof}.

% \begin{proof}[Sketch]
% For topological descriptors, by~\cite{skraba2020wasserstein,dlotko2023euler}, the $1$-Wasserstein distance between the Betti vectors is at most $k$, implying that the Betti numbers can change by at most the number of edge insertions or deletions affecting the underlying simplicial complex. For spectral measures, eigenvalues of the Laplacian matrix are Lipschitz continuous with respect to edge perturbations~\cite{von2007tutorial}, leading to a bounded Wasserstein distance proportional to the number of edge changes.
% \end{proof}

%% file: sections/04-experiments.tex
% This section shows the performance of T3former against some baseline models for temporal graph classification. Further ablation studies, hyperparameter analysis and robustness can be found in Appendix. \SC{Add appendix name}

% %\subsection{Temporal Graph Classification}

\subsection{Experimental Setup}
\label{sec:exp-setting}

\paragraph{Datasets.}
For the temporal graph classification experiments, we employed benchmarks across three domains: social, brain, and traffic networks. The social-network datasets consist of five temporal graphs from the TUDataset collection, \texttt{Infectious}, \texttt{DBLP}, \texttt{Tumblr}, \texttt{MIT}, and \texttt{Highschool}~\cite{morris2020tudataset}. The brain-network benchmarks are drawn from the NeuroGraph collection and include \texttt{DynHCP-Task}, \texttt{DynHCP-Gender}, and \texttt{DynHCP-Age}~\cite{said2023neurograph}. The traffic benchmarks are derived from the PEMS collection, \texttt{PEMS04}, \texttt{PEMS08}, and \texttt{PEMSBAY},~\cite{guo2019attention}. These traffic datasets were originally designed for temporal regression tasks; to enable classification, we reformulated them as binary and three-class temporal graph classification problems. Our adaptation extends their utility to new learning settings. Key dataset characteristics are summarized in~\Cref{tab:datasets}, with additional details on temporal structure and task definitions provided in~\Cref{app:datasets}.

\begin{table}[t]
\vspace{-.1in}
\caption{Temporal Graph Classification datasets. \label{tab:datasets}}
\centering
\resizebox{\linewidth}{!}{
\begin{tabular}{lcccccc}
\toprule
\textbf{Dataset} & \textbf{Graphs} & \textbf{Class} & \textbf{Avg.Node} & \textbf{Avg.Edge*} & \# \textbf{Node Feat.}& \textbf{timesteps} \\ 
\midrule
\texttt{Infectious}        & 200            & 2                & 50.00               & 459.72 & -- & 48      \\ 
\texttt{DBLP}             & 755         & 2        & 52.87               & 99.78   &--  & 46         \\ 
%Facebook          & 995               & 2               & 95.72         & 101.72 &-- &  104         \\ 
\texttt{Tumblr}            & 373               & 2                & 53.11    & 71.63  &--  & 89       \\ 
\texttt{MIT}               & 97               & 2                & 20     & 1469.15  &-- & 5576         \\ 
\texttt{Highschool}        & 180               & 2      & 52.32          & 544.81   &-- & 203     \\ 
\midrule
 \texttt{DynHCP-Task} & 7443& 7 &100 &843.04 &100 & 34 \\
  \texttt{DynHCP-Gender} & 1080& 2&100 &874.88 &100& 34 \\
  \texttt{DynHCP-Age}     &1067&3 &100 &875.42 &100&  34\\ 
  \midrule
 \texttt{PEMS04} & 708 & 2/3 & 307 & 680 & 3 & 24 \\
  \texttt{PEMS08} & 744& 2/3& 170 & 548 &3& 24 \\
  \texttt{PEMSBAY}     &2172 &2/3 &325 & 2694 &1&  24\\ 
\bottomrule
\end{tabular}}

{\scriptsize \textit{*Avg. Edges counts edge occurrences per snapshot (with repeats), not unique edges.}}
\end{table}

\paragraph{Task.}  We focus on temporal graph classification, where each graph $\mathcal{G}$ has timestamped edges capturing structural evolution over time. The label $y$ is fixed and time-invariant. The goal is to predict $y$ by leveraging temporal connectivity patterns and relevant domain-specific signals.
%In this work, we focus on the task of temporal graph classification, where each graph $\mathcal{G}$ is composed of edges annotated with timestamps, capturing the dynamic evolution of its structure over time. Despite the temporal nature of the graph, the corresponding label $y$ remains fixed and time-invariant. The objective is to accurately predict this label by leveraging the temporal connectivity patterns and other relevant domain-specific information embedded in the graph’s evolution.

\paragraph{Model Configuration.}  
We select the window length $\delta = 6$ and stride $\sigma = 4$ based on validation (\Cref{app:sliding-window}), resulting in $ N = \left\lceil \frac{t_{\max} - t_{\min} - \delta}{\sigma} \right\rceil + 1$
windows per temporal graph. From each window, we extract the following information:
\begin{itemize}
  \item \emph{Topological tokens:} Betti-0 and Betti-1 computed from the clique complex of each window.
  \item \emph{Spectral tokens:} Degree of Spectrality (DoS) histogram with 4 bins over the normalized Laplacian spectrum.
  \item \emph{Global encoder:} A 2-layer \textsc{GSAGE} encoder~\cite{hamilton2017inductive} is applied to the static graph, using either \textit{temporal degree} features or domain-specific node features.
\end{itemize}
% We perform hyperparameter tuning over the learning rate $\{0.01, 0.005, 0.001\}$, dropout rate $\{0.0, 0.3, 0.5\}$, and hidden dimensions $\{16, 32, 64, 128\}$. 
Topological and spectral tokens are processed through separate Transformer encoders, while the static graph with node features is encoded using \textsc{GSAGE}. To better exploit temporal information in node features, we replace the standard one-hot degree vectors with \textit{temporal degree vectors} as initial embeddings (\Cref{app:temporal-degree}). Each of the three branches produces a 10-dimensional temporal graph representation. These are concatenated into a 30-dimensional representation, which is then passed through a self-attention layer to adaptively fuse the information. The final attended representation is fed into a linear layer for classification.

\begin{table*}[t]
\centering
\caption{{\bf Social Networks.} Temporal graph classification accuracy results on social network datasets, following the Temporal Graph Benchmark~\cite{tieu2024temporal}.}
\label{tab:social}
\resizebox{0.8\textwidth}{!}{
\begin{tabular}{clccccc}
\toprule
& \textbf{Method}             & \textbf{Infectious}       & \textbf{DBLP}              & \textbf{Tumblr}            & \textbf{MIT}               & \textbf{Highschool}        \\
\midrule
\multirow{3}{*}{\textbf{Kernels}} 
& \textsc{Shortest Path}~\cite{borgwardt2005shortest} & $\pmval{67.00}{7.50}$ & $\pmval{56.00}{4.90}$      & $\pmval{58.00}{14.30}$     & $\pmval{50.80}{2.90}$      & $\pmval{56.00}{8.00}$      \\
& \textsc{Random Walk}~\cite{vishwanathan2010graph}    & $\pmval{67.00}{7.30}$ & $\pmval{53.00}{5.80}$      & $\pmval{58.00}{11.20}$     & $\pmval{56.80}{14.20}$     & $\pmval{53.74}{5.40}$      \\
& \textsc{WL-Subtree}~\cite{shervashidze2011weisfeiler} & $\pmval{60.00}{4.40}$ & $\pmval{52.00}{6.80}$      & $\pmval{57.00}{12.10}$     & $\pmval{55.50}{11.40}$     & $\pmval{55.20}{8.20}$      \\
\midrule
& \textsc{GSAGE}~\cite{hamilton2017inductive}    & $\pmval{58.00}{4.11}$ & $\pmval{56.29}{0.66}$ & $\pmval{59.78}{4.95}$ &\textcolor{blue}{${\pmval{72.00}{9.91}}$} & $\pmval{62.78}{7.51}$ \\
\textbf{Static} & \textsc{Graph2Vec}~\cite{narayanan2017graph2vec}    & $\pmval{56.50}{8.10}$ & $\pmval{53.90}{3.10}$      & $\pmval{54.70}{7.10}$      & $\pmval{52.20}{10.10}$     & $\pmval{57.60}{7.00}$      \\
\textbf{GNNs}& \textsc{NetLSD}~\cite{tsitsulin2018netlsd}             & $\pmval{62.50}{6.10}$ & $\pmval{55.80}{3.50}$      & $\pmval{55.20}{4.60}$      & $\pmval{60.10}{12.40}$     & $\pmval{55.20}{5.90}$      \\
 &
\textsc{GL2Vec}~\cite{chen2019gl2vec}                  & $\pmval{54.50}{5.10}$ & $\pmval{56.20}{3.00}$      & $\pmval{55.80}{8.00}$      & $\pmval{55.20}{4.80}$      & $\pmval{57.60}{4.10}$      \\

\midrule
 &\textsc{TGN}~\cite{rossi2020temporal}                 & $\pmval{52.00}{1.90}$ & $\pmval{58.00}{0.30}$      & $\pmval{51.70}{2.50}$      & $\pmval{60.23}{5.37}$      &\textcolor{blue}{${\pmval{63.33}{6.83}}$}     \\
\textbf{Temporal} & \textsc{EvolveGCN}~\cite{pareja2020evolvegcn}        & $\pmval{52.10}{9.30}$ & $\pmval{40.00}{8.90}$      & $\pmval{39.50}{8.90}$      & $\pmval{40.00}{9.45}$      & $\pmval{41.73}{9.32}$      \\
\textbf{GNNs}& \textsc{GraphMixer}~\cite{cong2023we}                 & $\pmval{50.00}{0.00}$ & $\pmval{56.30}{1.10}$      & $\pmval{50.90}{5.00}$      & $\pmval{56.28}{3.33}$                        & $\pmval{57.61}{7.88}$                        \\
& \textsc{Temp-G$^3$NTK}~\cite{tieu2024temporal}       &\textcolor{blue}{$\mathbf{\pmval{74.00}{5.80}}$} & \textcolor{blue}{${\pmval{60.00}{6.30}}$} &\textcolor{blue}{${\pmval{63.00}{6.80}}$} & $\pmval{58.40}{11.50}$     & $\pmval{58.10}{4.20}$      \\
\midrule
\textbf{Ours} &
\textsc{T3Former}                                       & \textcolor{blue}{${\pmval{68.50}{6.30}}$} &\textcolor{blue}{$\mathbf{\pmval{60.90}{0.70}}$} &\textcolor{blue}{$\mathbf{\pmval{63.20}{3.20}}$} & \textcolor{blue}{$\mathbf{\pmval{73.16}{4.13}}$} &\textcolor{blue}{$\mathbf{\pmval{67.20}{3.20}}$} \\
\bottomrule
\end{tabular}}
\vspace{-.15in}
\end{table*}

\begin{table}[t]
\centering
\caption{{\bf Traffic Networks.} Temporal graph classification accuracy results on traffic network datasets, evaluated under both binary and 3-class label settings.}
\label{tab:traffic}
\resizebox{\linewidth}{!}{
\begin{tabular}{lcccccc}
\toprule
 \multirow{2}{*}{\textbf{Method}} & \multicolumn{2}{c}{\textbf{PEMS04}} & \multicolumn{2}{c}{\textbf{PEMS08}} & \multicolumn{2}{c}{\textbf{PEMSBAY}} \\
\cmidrule(lr){2-3} \cmidrule(lr){4-5} \cmidrule(lr){6-7}
 & \textbf{Binary} & \textbf{Multi} & \textbf{Binary} & \textbf{Multi} & \textbf{Binary} & \textbf{Multi} \\
\midrule

\textsc{GSAGE} & $\pmval{85.31}{3.48}$  & $\pmval{80.79}{4.37}$  & $\pmval{93.41}{1.11}$  & $\pmval{85.62}{1.99}$& $\pmval{84.61}{1.74}$ &$\pmval{71.40}{6.10}$ \\
\textsc{GAT} & $\pmval{87.42}{6.32}$  & $\pmval{84.59}{5.20}$  & $\pmval{92.61}{2.37}$  & $\pmval{86.69}{3.78}$ &$\pmval{93.39}{0.43}$ &$\pmval{88.30}{3.93}$ \\
\textsc{UniMP} & $\pmval{86.73}{5.05}$  & $\pmval{80.93}{4.38}$  & $\pmval{93.28}{1.95}$  & $\pmval{85.49}{3.00}$&$\pmval{90.60 }{8.04}$ & $\pmval{73.51}{8.08}$ \\
\textsc{Graphormer}& \textcolor{blue}{$\pmval{93.64}{1.76}$}  & \textcolor{blue}{$\pmval{90.26}{5.48}$}  & \textcolor{blue}{$\pmval{93.82}{3.09}$}  & \textcolor{blue}{$\pmval{87.71}{2.89}$}&$\pmval{94.08}{0.49}$ &\textcolor{blue}{$\pmval{89.93}{1.49}$} \\

\midrule
\textsc{TGN}&$\pmval{86.30}{4.11}$ &$\pmval{80.09}{3.25}$ &$\pmval{90.73}{2.66}$ &$\pmval{83.87}{3.19}$ &\textcolor{blue}{$\pmval{95.44}{1.15}$} &$\pmval{88.85}{1.30}$ \\
\textsc{GCN+LSTM} &$\pmval{91.18}{3.26}$ &$\pmval{86.02}{1.61}$ &$\pmval{90.05}{1.86}$ &$\pmval{84.81}{2.77}$ &$\pmval{94.86}{2.21}$ &$\pmval{85.47}{2.52}$ \\
\textsc{TGAT} & $\pmval{90.85}{7.18}$& $\pmval{86.02}{3.18}$&$\pmval{82.77}{9.49}$ &$\pmval{87.23}{3.21}$ &$\pmval{94.08}{2.64}$ &$\pmval{88.75}{4.66}$ \\
\midrule
\textsc{T3Former} & \textcolor{blue}{$\mathbf{\pmval{96.76}{1.92}}$} & \textcolor{blue}{$\mathbf{\pmval{92.66}{1.93}}$} & \textcolor{blue}{$\mathbf{\pmval{95.16}{1.50}}$} & \textcolor{blue}{$\mathbf{\pmval{89.65}{2.02}}$}   &  \textcolor{blue}{$\bf\pmval{96.68}{1.30}$} & \textcolor{blue}{$\bf\pmval{92.35}{1.52}$} \\
\bottomrule
\end{tabular}
}
\vspace{-.15in}
\end{table}

\paragraph{Training Details.}  
We train our model using the Adam optimizer with a cross-entropy loss function and a weight decay of $1\mathrm{e}{-4}$. Hyperparameter tuning is performed via grid search over the following ranges: learning rate $\{0.01, 0.005, 0.001\}$, dropout rate $\{0.0, 0.3, 0.5\}$, and hidden dimensions $\{16, 32, 64, 128\}$.
For the social and traffic networks, we follow a 5-fold cross-validation protocol consistent with the experimental setup in~\cite{tieu2024temporal}. For the brain network dataset, we adopt a 70/10/20 train/validation/test split using the same random seed as the baseline paper~\cite{said2023neurograph}. Our implementation is written in Python and experiments are conducted on a server equipped with an NVIDIA H100 NVL GPU and 768 GB of RAM. %The code is publicly available at: \url{https://anonymous.4open.science/r/T3Former-3311/}.

\paragraph{Computational Complexity and Runtime.} Let $n=|\mathcal{V}|$, $m=|\mathcal{E}|$, and 
$N=\Bigl\lceil\frac{t_{\max}-t_{\min}-\delta}{\sigma}\Bigr\rceil+1$
be the number of sliding windows. Temporal filtration takes $O(m)$ per window (total $O(Nm)$), topological descriptors cost $C_{\mathrm{topo}}(n,m)$ per window (worst‑case $O(n^3)$, typically $O(m^\omega)$ for sparse graphs where $\omega<2.4$~\cite{otter2017roadmap,milosavljevic2011zigzag}), and spectral DOS extraction with $k$ Lanczos/Chebyshev steps requires $O(km)$ per window~\cite{weisse2006kernel}. Running \textsc{GSAGE} with $L$ layers and hidden dimension $d$ on the static graph costs $O(Lmd)$, and fusing the DoS and Betti sequences of length $N$ via self‑attention incurs $O(N^2d + Nd^2)$. Hence, the overall complexity is  

$\mathcal{O}\bigl(N\,(m + C_{\mathrm{topo}}(n,m) + km)\;+\;Lmd\;+\;(N^2d + Nd^2)\bigr),$

\noindent which remains practical for moderate $n$, $m$, and $N$. %\BC{Joshem, add total runtime for one large dataset with the hardware details.}

Our model demonstrates a clear computational advantage by efficiently integrating temporal information without the need for separate static GNN processing followed by sequential learning. This streamlined design results in significantly reduced runtime. On the large-scale \texttt{Task} dataset, a complete run of our model takes approximately $98$ minutes. More notably, for a single fold, our model requires only $5.34$ minutes, whereas the GCN+LSTM baseline takes $84.42$ minutes under the same configuration, making our approach roughly 15 times faster. The detailed runtime comparison is presented in~\Cref{tab:runtime}, highlighting the efficiency and scalability of our method.

\subsection{Results} 

\paragraph{Baselines.}
We evaluate \textsc{T3Former} across three domains: social, brain, and traffic networks. For \textbf{social networks}, we adopt the Temporal Graph Benchmark~\cite{tieu2024temporal}. Comparison models include (i) \textit{Graph kernel methods} such as \textsc{WL-Subtree}, \textsc{Shortest-Path}, and \textsc{Random-Walk}, which respectively capture neighborhood structures, path-length distributions, and walk-based similarities; (ii) \textit{Static GNNs} including \textsc{GSAGE} (inductive message passing), \textsc{Graph2Vec} (skip-gram on rooted subgraphs), \textsc{NetLSD} (spectral heat trace descriptors), and \textsc{GL2Vec} (graphlet-based features); and (iii) \textit{Temporal GNNs} such as \textsc{GraphMixer}, \textsc{TGN}, \textsc{EvolveGCN}, and \textsc{Temp-G$^3$NTK}, which incorporate temporal dynamics through attention, memory modules, or time-aware kernels.
For \textbf{brain networks}, we follow the NeuroGraph benchmark~\cite{said2023neurograph}, comparing against \textsc{UniMP}, \textsc{k-GNN}, \textsc{GAT}, \textsc{GSAGE}, and \textsc{Gen-GNN}. Relevant references are provided in~\Cref{tab:social,tab:brain}. In \textbf{traffic networks}, we additionally evaluate against the transformer-based \textsc{Graphormer}~\cite{ying2021transformers}, as well as temporal baselines including \textsc{TGN}~\cite{rossi2020temporal}, \textsc{GCN+LSTM}~\cite{yu2018spatio}, and \textsc{TGAT}~\cite{xu2020inductive}.

\paragraph{Social Network Results.} \Cref{tab:social} compares \textsc{T3Former} against classical graph kernels, static GNNs, and recent temporal GNNs on five temporal social graphs.  Graph kernels and static GNNs achieve only moderate accuracy and exhibit wide variations across datasets, while temporal GNNs improve on some networks but remain uneven in others.  In contrast, \textsc{T3Former} delivers consistently strong performance on every benchmark, outperforming most baselines on each dataset and narrowing the gap to the very best method where it does not lead outright.  This uniform gain highlights the strength of our hybrid design: by fusing timestamp‑aware topological signatures with spectral descriptors through self‑attention, \textsc{T3Former} robustly captures both local dynamics and global structure in evolving graphs.

%This consistent gain underscores the strength of our hybrid design: by fusing timestamp-aware topological and spectral descriptors via self-attention, \textsc{T3Former} effectively captures both local dynamics and global structure in evolving graphs.

\begin{table}[t]
  \centering
\caption{{\bf Brain Networks.} Temporal graph classification accuracy results on brain networks, following NeuroGraph benchmark~\cite{said2023neurograph}.}
  \label{tab:brain}
  \resizebox{.8\linewidth}{!}{
  \begin{tabular}{lccc}
    \toprule
    \textbf{Method} & \textbf{Task} & \textbf{Gender} & \textbf{Age} \\
    \midrule
        \textsc{GSAGE}~\cite{hamilton2017inductive}          & \textcolor{blue}{\textbf{90.93}} & 66.20          & 40.65 \\
            \textsc{GAT}~\cite{velivckovic2017graph}            & 89.67          & 67.13          & 44.39 \\
                \textsc{k--GNN}~\cite{morris2019weisfeiler}         & 73.03          & 68.45          & 44.25 \\
    \textsc{UniMP}~\cite{shi2020masked}         & 89.66          & \textcolor{blue}{72.30} & \textcolor{blue}{44.41} \\
    \textsc{Gen-GNN}~\cite{you2020design}        & 68.84          & 62.04          & 42.99 \\
    \midrule
    \textsc{T3former}       & \textcolor{blue}{90.76}            & \textcolor{blue}{\textbf{75.79}}            & \textcolor{blue}{\textbf{58.73}}    \\
    \bottomrule
  \end{tabular}}
  \vspace{-.15in}
\end{table}

\begin{table*}[t]
\centering
\caption{{\bf Ablation Study.} Comparison of the standalone performance of each component in our architecture. Concat‑fuse represents concatenation of these features instead of attention‑based integration.}
\label{tab:ablation}
\resizebox{\linewidth}{!}{
\begin{tabular}{lccccc|ccc|cccccc}
\toprule
& \multicolumn{5}{c}{\textbf{Social Networks}} & \multicolumn{3}{c}{\textbf{Brain Networks}} & \multicolumn{6}{c}{\textbf{Traffic Networks}} \\
\cmidrule(lr){2-6} \cmidrule(lr){7-9} \cmidrule(lr){10-15}
\textbf{Method} & \textbf{Infect.} & \textbf{DBLP} & \textbf{Tumblr} & \textbf{MIT} & \textbf{HSchool} & \textbf{Task} & \textbf{Gender} & \textbf{Age} & \textbf{P04-2} & \textbf{P04-3} & \textbf{P08-2} & \textbf{PS08-3} & \textbf{PBAY-2} & \textbf{PBAY-3} \\
\midrule
\textsc{GSage}        & 58.00   & 56.29   & 59.78   & 72.00   & 62.78   & \textcolor{blue}{\textbf{91.57}} & 63.83   & 37.55   & 85.31   & 80.79   & 93.41   & 85.62   & 84.61   & 71.40   \\
\textsc{Topo+Tr}      & 63.50 & 58.60   & 61.70 & 61.90   & 63.90   & 22.10         & 58.80   & 45.64   & 85.03   & 69.63   & 80.11   & 71.51   & 94.54   & 81.22   \\
\textsc{DoS+Tr}       & \textcolor{blue}{65.00}   & \textcolor{blue}{59.80} & 61.90   & 70.10   & 65.00   & 23.31         & 57.87   & 45.83   & 82.49   & 67.66   & 78.09   & 68.55   & 89.46   & 76.61   \\
\textsc{Concat‑Fuse}  & 63.50   & 58.94   & \textcolor{blue}{63.00}   & \textcolor{blue}{72.16} & \textcolor{blue}{65.56} & 90.58         & \textcolor{blue}{73.58} & \textcolor{blue}{57.78} & \textcolor{blue}{95.90} & \textcolor{blue}{92.52} & \textcolor{blue}{95.03} & \textcolor{blue}{\textbf{89.92}} & \textcolor{blue}{\textbf{96.82}} & \textcolor{blue}{90.47} \\
\midrule
\textsc{T3Former}     & \textcolor{blue}{\textbf{68.50}} & \textcolor{blue}{\textbf{60.90}} & \textcolor{blue}{\textbf{63.20}} & \textcolor{blue}{\textbf{73.16}} & \textcolor{blue}{\textbf{67.20}} & \textcolor{blue}{90.76} & \textcolor{blue}{\textbf{75.79}} & \textcolor{blue}{\textbf{58.73}} & \textcolor{blue}{\textbf{96.76}} & \textcolor{blue}{\textbf{92.66}} & \textcolor{blue}{\textbf{95.16}} & \textcolor{blue}{89.65} & \textcolor{blue}{96.68} & \textcolor{blue}{\textbf{92.35}} \\
\bottomrule
\end{tabular}
}
\end{table*}

% \begin{table*}[t]
% \centering
% \caption{{\bf Ablation Study.} Comparison of the standalone performance of each component in our architecture. Concat-fuse represents concatenation of these features instead of attention based integration.}
% \label{tab:ablation}
% \resizebox{\linewidth}{!}{
% \begin{tabular}{lccccc|ccc|cccccc}
% \toprule
% & \multicolumn{5}{c}{\textbf{Social Networks}} & \multicolumn{3}{c}{\textbf{Brain Networks}} & \multicolumn{6}{c}{\textbf{Traffic Networks}} \\
% \cmidrule(lr){2-6} \cmidrule(lr){7-9} \cmidrule(lr){10-15}
% \textbf{Method} & \textbf{Infect.} & \textbf{DBLP} & \textbf{Tumblr} & \textbf{MIT} & \textbf{HSchool} & \textbf{Task} & \textbf{Gender} & \textbf{Age} & \textbf{P04-2} & \textbf{P04-3} & \textbf{P08-2} & \textbf{PS08-3} & \textbf{PBAY-2} & \textbf{PBAY-3} \\
% \midrule
% \textsc{GSage}     & 58.00 & 56.29 & 59.78 & 72.00 & 62.78 & 91.57 & 63.83 & 37.55 & 85.31 & 80.79 & 93.41 & 85.62 & 84.61 & 71.40 \\
% \textsc{Topo+Tr}   & 66.50 & 58.60 & 63.10 & 61.90 & 63.90 & 22.10 & 58.80 & 45.64 & 85.03 & 69.63 & 80.11 & 71.51 & 94.54 & 81.22 \\
% \textsc{DoS+Tr}   & 64.50 & 59.80 & 60.30 & 70.10 & 65.00 & 23.31 & 57.87 & 45.83 & 82.49 & 67.66 & 78.09 & 68.55 & 89.46 & 76.61 \\
% \textsc{Concat-Fuse} & 63.50&58.94 &63.00 & 72.16 &65.56 & 90.58&73.58 &57.78 & 95.90& 92.52&95.03 &89.92 &96.82&90.47  \\
% \midrule
% \textsc{T3Former}  & 68.50 & 60.90 & 63.20 & 73.16 & 67.20 & 90.76 & 75.79 & 58.73 & 96.76 & 92.66 & 95.16 & 89.65 & 96.68 & 92.35 \\
% \bottomrule
% \end{tabular}
% }
% \end{table*}

\begin{table*}[t]
\centering
\caption{\textbf{Attention Allocation.} Total attention weight assigned to each \textsc{T3former} component, GSAGE (structural), Topological, and DoS (spectral), for each dataset across the social, brain, and traffic network benchmarks.}
\label{tab:attention}
\resizebox{\linewidth}{!}{
\begin{tabular}{lccccc|ccc|cccccc}
\toprule
& \multicolumn{5}{c}{\textbf{Social Networks}} & \multicolumn{3}{c}{\textbf{Brain Networks}} & \multicolumn{6}{c}{\textbf{Traffic Networks}} \\
\cmidrule(lr){2-6} \cmidrule(lr){7-9} \cmidrule(lr){10-15}
\textbf{Method} & \textbf{Infect.} & \textbf{DBLP} & \textbf{Tumblr} & \textbf{MIT} & \textbf{HSchool} & \textbf{Task} & \textbf{Gender} & \textbf{Age} & \textbf{P04-2} & \textbf{P04-3} & \textbf{P08-2} & \textbf{P08-3} & \textbf{PBAY-2} & \textbf{PBAY-3} \\
\midrule
\textsc{GSage}    &0.3435 &0.8292 &0.9336 &0.8549 &0.8567 & 0.7040&0.3907 & 0.4691&0.3855 &0.6687 &0.5163 & 0.5920 &0.5645& 0.4419 \\
\textsc{Topo}  &0.3126 &0.1006 &0.0274 &0.1443 &0.0958 &0.1639 & 0.4557& 0.0635& 0.2679&0.1172 &0.3049 & 0.0816 &0.3873&0.4971  \\
\textsc{DoS}   &0.3439 &0.0702 &0.0390 &0.0009 &0.0475 &0.1321 & 0.1535& 0.4674&0.3466&0.2141 & 0.1789&0.3263 &0.0482 & 0.0610\\
\bottomrule
\end{tabular}
}
\vspace{-.2in}
\end{table*}

%As shown in~\Cref{tab:social}, classical graph kernels and static GNNs achieve only moderate and often uneven results across the five social‑network benchmarks, and even state‑of‑the‑art temporal GNNs can excel on one dataset but falter on others. By contrast, \textit{T3Former} delivers uniformly strong performance everywhere, consistently matching or outperforming the best baseline on each graph. This robustness underscores the value of our hybrid scheme, sliding‑window temporal filtration coupled with complementary topological and spectral descriptors, all fused via self‑attention, which captures both detailed temporal dynamics and overarching structural patterns far more effectively than any single prior approach.

\paragraph{Traffic Network Results.} 
The temporal graph classification accuracy across three benchmark datasets, \texttt{PEMS04}, \texttt{PEMS08}, and \texttt{PEMSBAY}, under both binary and 3-class label settings is presented in Table~\ref{tab:traffic}. Our proposed model, \textsc{T3Former}, consistently outperforms all baseline methods across all settings. In particular, \textsc{T3Former} achieves the highest accuracy on every task, with an average gain of {1.90\%} and {2.25\%} over the best-performing baseline for the binary and multi-class settings, respectively. Among the baselines, the transformer-based \textsc{Graphormer} and inductive GNNs such as \textsc{GSAGE} and \textsc{GAT} perform competitively but fall short in capturing fine-grained temporal dynamics. For a fair comparison, static GNN baselines are evaluated on concatenated node features across all time steps, effectively flattening the temporal structure. While some temporal models, such as \textsc{GCN+LSTM} and \textsc{TGN}, show improvements over static counterparts, they still underperform compared to \textsc{T3Former}, particularly on more complex multi-class tasks. These results highlight the strength of our multi-view designed model.

\paragraph{Brain Connectivity Results.} 
\Cref{tab:brain} presents our results on the three DynHCP tasks alongside the NeuroGraph benchmarks~\cite{said2023neurograph}.  \textsc{T3Former}  delivers competitive performance compared to the top-performing method on the task detection, illustrating its ability to capture global temporal patterns in brain connectivity.  It outperforms all baselines on gender classification and achieves a substantial gain on age prediction, where modeling fine-grained, evolving connectivity is particularly challenging.  These improvements, together with the strong performance observed on social and traffic benchmarks, underscore the adaptability and versatility of \textsc{T3Former}, which consistently delivers superior results across diverse temporal‑graph domains.

\subsection{Ablation Studies}  
To dissect the contributions of each component in \textsc{T3Former}, we evaluate several simplified variants as summarized in Table~\ref{tab:ablation}. Specifically, we examine a static baseline (\textsc{GSAGE}), which collapses temporal dynamics into a single static embedding, as well as unimodal variants \textsc{Topo-TR} and \textsc{DoS-TR}, each employing a transformer-based architecture focused solely on topological or spectral descriptors, respectively. We also include a non-attentive fusion approach, \textsc{Concat-Fuse}, which concatenates static, topological, and spectral embeddings into a single representation without adaptive weighting.

Results highlight clear trade-offs among these modalities: while unimodal variants (\textsc{Topo-TR}, \textsc{DoS-TR}) perform competitively on datasets dominated by temporal connectivity (e.g., \texttt{Infectious}, \texttt{Tumblr}), they exhibit weaker performance in scenarios dominated by static or feature-driven information (e.g., brain network \texttt{Task}). Conversely, the static \textsc{GSAGE} baseline excels in such feature-rich contexts but lacks the expressive power to capture detailed temporal structure.

This complementary nature strongly motivates the adaptive cross-modal integration provided by \textsc{T3Former}'s Descriptor-Attention mechanism. As shown in Table~\ref{tab:attention}, attention allocation varies significantly across datasets, structural embeddings dominate social network tasks, whereas topological and spectral descriptors receive higher weights in certain traffic and brain network scenarios. Such adaptive weighting enables \textsc{T3Former} to flexibly harness the most relevant features, consistently delivering superior performance, robustness, and enhanced representation quality, as also validated through improved class separation in t-SNE visualizations (Appendix~\ref{app:tsne}).

%% file: sections/05-appendix.tex
\appendix

\section{Proofs of Stability Theorem} \label{sec:proof}

In this part, we provide complete proofs of the stability theorem stated in \Cref{sec:stability}.

\paragraph{Notation and Setup.}  
Let $\mathcal{G} = (\mathcal{V}, \mathcal{E}, \tau)$ be a temporal graph, where $\mathcal{V}$ is the set of nodes, $\mathcal{E}$ is the set of edges, and $\tau : \mathcal{E} \to \mathbb{R}^+$ is a \textit{function} which assigns a unique timestamp to each edge. Given a window $[t, t+\delta]$, the induced subgraph $\G_{[t, t+\delta]}$ consists of all edges whose timestamps fall within $[t, t+\delta]$, and the nodes incident to them. 

%For each $G_{[t, t+\delta]}$, we define two types of descriptors:
% \begin{itemize}
%     \item The \textit{topological descriptor} $\phi_t$ consists of graph statistics such as the number of nodes, number of edges, and Betti numbers $\beta_0$ and $\beta_1$ computed from the clique complex.
%     \item The \textit{spectral descriptor} $\psi_t$ is the histogram of the eigenvalues of the normalized Laplacian of $G_{[t, t+\delta]}$, approximating the density of states (DOS).
% \end{itemize}

% We consider perturbations where up to $k$ edge insertions or deletions are made within $G_{[t, t+\delta]}$, resulting in a modified subgraph $G'_{[t, t+\delta]}$ with corresponding descriptors $\phi'_t$ and $\psi'_t$.

\paragraph{Stability of Topological Descriptors.} In this part, we establish the stability of our topological descriptors. We first introduce a few key lemmas from algebraic topology. Throughout the following, $|\cdot|_p$ denotes the $L^p$-norm, and $\W_p(\cdot,\cdot)$ denotes the Wasserstein distance between persistence diagrams~\cite{dey2022computational}. The first lemma shows that a small change in the filtration function results in small change in the induced persistence diagrams.

\begin{lemma} \cite{skraba2020wasserstein} \label{lem:filtration} Let $\X$ be a compact metric space, and $f_1,f_2:\X\to\R$ be two filtration functions. Let $\PD_k(\X,f_i)$ be the $k^{th}$ persistence diagram for sublevel filtration via $f_i$. Then, for any $p\geq 1$, we have 
$$\W_p(\PD_k(\X,f_1),\PD_k(\X,f_2))\leq \|f_1-f_2\|_p$$
\end{lemma}

\smallskip

The next lemma is on the $L^1$-stability of Betti curves by~\cite{dlotko2023euler}.   

\begin{lemma} \cite{dlotko2023euler} \label{lem:betti} Let $\beta_k(\X)$ is the $k^{th}$ Betti function obtained from the persistence module $\PM_k(\X)$.
$$\|\vec{\beta}_k(\X)-\vec{\beta}_k(\Y)\|_1\leq 2\W_1(\PD_k(\X),\PD_k(\Y))$$
where $\vec{\beta}_k(\X)$ represents the Betti vectors corresponding to $\PD_k(\X)$.
\end{lemma}

\smallskip

Now, we are ready to prove our stability theorem.

\noindent {\bf \Cref{thm:stability-topo}.} \textit{Let $\G=(\V,\E) $ be a graph and let $\tau_1,\tau_2 :\E\to\R$ be two timestamp functions on $\G$. Then, for $k\geq 0$, we have 
$$\|\vec{\beta}_k(\G,\tau_1)-\vec{\beta}_k(\G,\tau_2)\|_1 \leq C_k\cdot \|\tau_1-\tau_2\|_1$$
where $\vec{\beta}_k(\G,\tau_i)$ represents the Betti vector corresponding to $\PD_k(\G,\tau_i)$ obtained by sublevel filtration with respect to $\tau_i$.}

%\BC{I'll modify this proof.}

\begin{proof} We will utilize the stability lemmas given above. Note that our result is more general and applies to any sliding window $\G_{[t,t+\delta]}$. Let $\wh{\G}$ be the clique complex of $\G$. Then, by using the filtration functions $\tau_1,\tau_2:\E\to\R$, we can induce a continuous function on the metric graph $\wt{\tau}_1,\wt{\tau_2}:\G\to\R$ with $\max_{x\in e}\wh{\tau_i}(x)= \{\tau_i(e)\}$ and $\wt{\tau}_i(v)=\min_{e_j\supset v}\tau_i(e_j)$. Then, we construct continuous extensions of $\wt{\tau}_i$ on the clique complex by linear extensions in each simplex, i.e., $\wh{\tau}_1,\wh{\tau_2}:\wh{\G}\to\R$.

Then, by Lemma \ref{lem:filtration}, we have 
\begin{equation} \label{eqn1}
\W_1(\PD_k(\wh{\G},\wh{\tau}_1),\PD_k(\wh{\G},\wh{\tau}_2))\leq \|\wh{\tau}_1-\wh{\tau}_2\|_1\leq C^k_\G\cdot\|\tau_1-\tau_2\|_1
\end{equation}
% Recall that our Betti vectors $\vec{\beta}_k(\G,f)=[\beta_k(\wh{\G}_1) \  \beta_k(\wh{\G}_2) \  \dots \ \beta_k(\wh{\G}_{N-m})]$ where $\G_j$ is the induced subgraph from $\E_j=\{e\in\E\mid \tau_i(e)\leq \alpha_j\}$. 

% Let $\tau_1^j,\tau_2^j$ be the restriction of $\tau_i$ to $\E_j$. Then, by direct application of \Cref{eqn1}, we have 
% \begin{equation} \label{eqn2}
% \W_1(\PD_k(\wh{\G},\wh{\tau}),\PD_k(\wh{\G}_i,g_i))\leq \|f_i-g_i\|_1 \leq \|f-g\|_1
% \end{equation}
Note that the second inequality follows from $\wh{\tau}_i$ being the continuous extension of $\wt{\tau}_i$ on $\wh{\G}$. Next, by Lemma \ref{lem:betti},  we have 
\begin{equation} \label{eqn3}
\| \vec{\beta}_k(\wh{\G},\wh{\tau}_1)-\vec{\beta}_k(\wh{\G},\wh{\tau}_2)\|_1\leq 2\W_1(\PD_k(\wh{\G},\wh{\tau}_1),\PD_k(\wh{\G},\wh{\tau}_2))
\end{equation}

Hence, by combining \Cref{eqn1,eqn3}, we have 
\begin{align*}
 &\|\vec{\beta}_k(\G,\wh{\tau}_1)-\vec{\beta}_k(\G,\tau_2)\|_1  \leq 2C^k_\G\cdot \|\tau_1-\tau_2\|_1   
\end{align*}

The proof follows.
\end{proof}

\paragraph{Stability of Spectral Descriptors.} Next, we will prove the stability of density of states vectors under edge modifications.

\noindent {\bf \Cref{thm:stability-dos}.}
\textit{Let $\mathcal{G}$ and $\mathcal{G}'$ be two temporal graphs differing by at most $k$ edge modifications (insertions or deletions) within any window $[t, t+\delta]$. Then, the Wasserstein distance between the spectral descriptors $\psi_t$ and $\psi'_t$ is bounded by $Ck/n$, where $n = |\mathcal{V}|$ and $C$ is a constant depending on the eigenvalue distribution.}

%\BC{I'll revise the proof.}

\begin{proof}
Let $\mathcal{G}$ and $\mathcal{G}'$ be two temporal graphs differing by at most $k$ edge insertions or deletions within any window $[t, t+\delta]$. Let $\psi_t$ be the spectral descriptor for $\G_{[t,t+\delta]}$.

Next, we recall the following results from spectral graph theory. 
\begin{itemize}
    \item Adding or deleting an edge modifies at most two entries in the normalized Laplacian matrix.
    \item The matrix perturbation theory for eigenvalues (see, e.g.,~\cite{von2007tutorial}) implies that if $L$ and $L'$ differ by a matrix of spectral norm at most $\epsilon$, then the eigenvalues of $L$ and $L'$ differ by at most $\epsilon$.
    \item Since each edge modification changes at most $O(1/n)$ in spectral norm~\cite{chung1997spectral}, $k$ edge changes induce a perturbation of size at most $O(k/n)$.
    \item Therefore, the Wasserstein (matching) distance between the DOS histograms is bounded as
    $    \W_1(\psi_t, \psi'_t) = O\left(\frac{k}{n}\right)$,   
    where $\W_1$ denotes the 1-Wasserstein distance.
\end{itemize}
This implies $    \W_1(\psi_t, \psi'_t) \leq Ck/n$ for some $C>0$. The proof follows.
\end{proof}

% \begin{figure}[h!]
% \centering
% \includegraphics[width=.8\linewidth]{figures/graph-filtration.png}
% \caption{\footnotesize \textbf{Graph Filtration.} For $\mathcal{G}=\mathcal{G}_3$ in both examples, the top figure illustrates a \textit{superlevel filtration using the node degree function} with thresholds $3>2>1$, where nodes of degree 3 are activated first, followed by those of lower degrees.  Similarly, the bottom figure illustrates a \textit{sublevel filtration based on edge weights} with thresholds $1.5< 1.8< 2.1$.  \label{fig:graph-filtration}}
% \end{figure}

\begin{figure*}[t]
    \centering
    \begin{subfigure}[b]{0.3\textwidth}
        \centering
        \includegraphics[width=\textwidth]{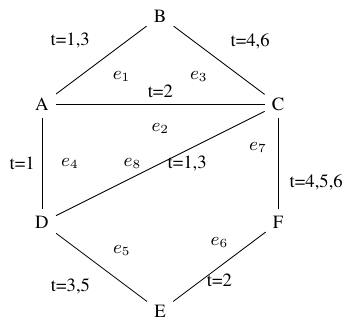}
        \vspace{.3in}
        \caption{Temporal Graph}
        \label{fig:TG}
    \end{subfigure}
    \hfill
    \begin{subfigure}[b]{0.65\textwidth}
        \centering
        \includegraphics[width=\textwidth]{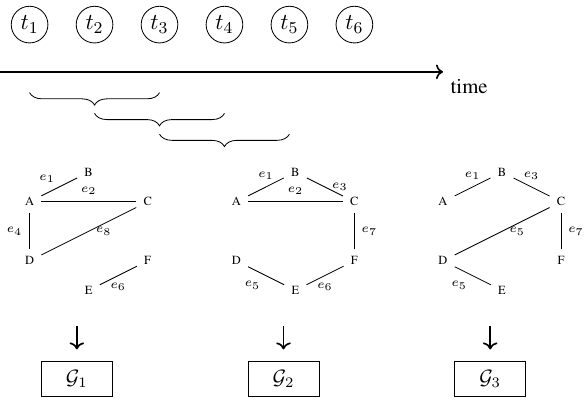}
        \caption{Sliding Window Subgraphs}
        \label{fig:SW}
    \end{subfigure}
    \caption{\textbf{Sliding‐window Subgraphs:} Given a temporal graph $\G$, the sliding window approach generates a sequence of subgraphs ($\G_{[1,3]}, \G_{[2,4]}, \G_{[3,5]}$ above), where each subgraph $\G_{[a,b]}$ includes only the edges active within the time interval $[a,b]$, i.e., edges $e$ such that $t_e \in [a,b]$.}
    % \caption{For a given temporal graph $\G$, sliding window approach induce a sequence of subgraphs $\G_{[1,3]},\G_{[2,4]}, \G_{[3,5]}$, by retaining only those edges $e$ whose activity time $t_e$ lies in the window $[a,b]$.}
    \label{fig:temporal}
\end{figure*}

\section{Experimental Details} \label{app:experiments}

\subsection{Datasets} \label{app:datasets}

\paragraph{TUDataset Temporal Graph Benchmarks.}  
We evaluate on six real-world temporal-graph classification datasets drawn from the TUDataset collection~\cite{morris2020tudataset}:  
\begin{itemize}[nosep,left=1em]
  \item \texttt{INFECTIOUS}: person‐to‐person proximity during an outbreak scenario.  
  \item \texttt{DBLP}: author‐collaboration evolution over time.  
  %\item \texttt{FACEBOOK}: link‐formation events among users.  
  \item \texttt{TUMBLR}: posting and reblogging interactions in a social network.  
  \item \texttt{MIT}: proximity traces from the Reality Mining study at MIT.  
  \item \texttt{HIGHSCHOOL}: face‐to‐face contacts recorded at a French high school.  
\end{itemize}  
All contact or interaction events are aggregated into fixed‐length windows (e.g.\ 5 min), producing sequences of graphs with a shared node set and time-varying edges.  

% The social network datasets did not have node features, so we created temporal degree features as the initial embedding for GSAGE. Temporal degree assigns a feature vector to every node corresponding to the degree of the node at every time step. For example, in \Cref{fig:temporal}, the temporal degree vector for node $A$ would be $[2,1,1,0,0,0]$, and the temporal degree vector for node $B$ would be $[1,0,1,1,0,1]$. Brain network and traffic network datasets had node features that allowed us to use domain information but in the absence of node features, our model creates meaningful feature vectors that improve temporal graph learning.

\paragraph{Dynamic HCP Brain Networks (DynHCP).}  
Following Said et al.~\cite{said2023neurograph}, we also apply our model to three splits of the Human Connectome Project task‐based fMRI data:  
\begin{itemize}[nosep,left=1em]
  \item \texttt{DynHCP-Task}: labeled by cognitive task (e.g.\ language, motor).  
  \item \texttt{DynHCP-Gender}: labeled by subject gender (male vs.\ female).  
  \item \texttt{DynHCP-Age}: labeled by age cohort (younger vs.\ older adults).  
\end{itemize}  

For DynHCP datasets, Each graph snapshot is generated using a 30-second sliding window over 200 cortical regions, with edge weights defined by the Pearson correlation between regional time series. A window of 50 timepoints is slid with a stride of 3 over a 150-length ROI time series, producing $\frac{150 - 50}{3} + 1 = 34$ snapshots per subject (i.e., 34 timesteps) .

\paragraph{Traffic Network Datasets.}
All traffic datasets are sourced from the California Department of Transportation’s Performance Measurement System (PeMS)~\cite{chen2001freeway} and are aggregated at 5‑minute intervals~\cite{guo2019attention}. We model each dataset as a graph $G=(V,E)$, where $V$ is the set of loop‑detector sensors and $E$ encodes pairwise road‑network distances. At each time step, the recorded variables include traffic flow (vehicle count), speed (km/h), and occupancy (\%).

\begin{itemize}
    
\item \texttt{PEMS04} consists of 307 sensors in the San Francisco Bay Area, covering the period from January 1, 2018, to February 28, 2018. It comprises 16,992 time steps. Each snapshot includes three channels (flow, speed, occupancy), with no missing values. 

\item \texttt{PEMS08} contains 170 highway sensors in the San Bernardino region, spanning July 1, 2016 to August 31, 2016 (17,856 time steps). As with PEMS04, we record flow, speed, and occupancy at 5‑minute intervals without gaps. 

\item \texttt{PEMS‑BAY} covers 325 sensors across the greater Bay Area, from January 1, 2017 to May 31, 2017 (52,128 time steps). This dataset provides only speed measurements, with a missing‐rate of approximately 0.02 \%. 
\end{itemize}

\begin{table*}[t]
\centering
\caption{\textbf{Runtime.} Total runtime for \textsc{T3Former} including Topological and DoS feature extraction, and attention-based classification for each dataset across the social, brain, and traffic network benchmarks.}
\label{tab:runtime}
\resizebox{\linewidth}{!}{
\begin{tabular}{lccccc|ccc|cccccc}
\toprule
& \multicolumn{5}{c}{\textbf{Social Networks}} & \multicolumn{3}{c}{\textbf{Brain Networks}} & \multicolumn{6}{c}{\textbf{Traffic Networks}} \\
\cmidrule(lr){2-6} \cmidrule(lr){7-9} \cmidrule(lr){10-15}
\textbf{Features} & \textbf{Infect.} & \textbf{DBLP} & \textbf{Tumblr} & \textbf{MIT} & \textbf{HSchool} & \textbf{Task} & \textbf{Gender} & \textbf{Age} & \textbf{P04-2} & \textbf{P04-3} & \textbf{P08-2} & \textbf{P08-3} & \textbf{PBAY-2} & \textbf{PBAY-3} \\
\midrule
\textsc{Topo}  & 1.88 & 6.89 & 2.52 & 1.63 & 0.66 & 3013.36 & 692.31 & 961.26 & 91.47 & 91.47 &54.83 & 54.83 &414.31 & 414.31 \\
\textsc{DoS}  &6.71 &33.45 &16.8 &9.81 &3.26 & 1412.33 & 157.43& 157.84& 193.44& 193.44 & 110.88 & 110.88 &211.39 & 211.39  \\
%\textsc{Temp. Degree}   & 7.29 &23.20 &15.25 &21.33 &88.18 & - & - & - &- &- & - & - & - & - \\
\midrule 
\textsc{T3Former} &8.77 &33.12 &18.18 & 8.56 &9.73 &1486.50 &508.50 &279.34 &35.71 &58.74 &36.48 &38.84 &105.00&108.00 \\
\midrule
\textsc{Total} & 24.65 & 96.66 & 52.75 & 41.33 & 110.83 & 5912.19 & 1358.24 & 1398.44 & 320.62 & 343.65 & 202.19 & 204.55 & 730.70 & 733.70 \\
\bottomrule
\end{tabular}
}
%\vspace{-.2in}
\end{table*}

\paragraph{Temporal Graph Classification with Traffic Networks.}
Although the PEMS datasets are traditionally used for temporal graph regression tasks, we reformulate them for temporal graph classification to establish a benchmark in this setting. In these datasets, each node represents a traffic sensor on a road, and edges are defined by intersections between roads. To construct temporal graphs, we partition the traffic data into 2-hour intervals and compute the mean link speed for each interval as $\overline{v}_d = \frac{1}{|\mathcal{E}|}\sum_{e \in \mathcal{E}} v_{e,d},$
where $v_{e,d}$ denotes the average speed on edge $e$ during interval $d$. For binary classification, we assign a label of \textit{congested} to the bottom 35\% of link speed intervals and \textit{free-flowing} to the remaining 65\%. For three-class classification, we define the classes as follows: the bottom 35\% as \textit{congested}, the next 25\% (35–60\%) as \textit{mildly congested}, and the top 40\% as \textit{free-flowing}. The resulting temporal graph classification datasets are publicly available with our code.

% \paragraph{Temporal Graph Classification with Traffic Networks.} While PeMS datasets are classically used for regression tasks, we reformulated them into temporal classification datasets to create a benchmark in this domain. A node in this dataset is a sensor on a road and if there is an intersection between two roads, there exists an edge between the nodes. In order to create temporal graphs, we divide the traffic data into $2-$hour intervals. Then, we compute mean link speed for each interval: \[\overline{v}_d = \frac{1}{|\mathcal{E}|}\sum_{e\in \mathcal{E}}v_{e,d}\] where $v_{e,d}$ is the average speed on edge $e$. Then, in order to create binary labels, we label the lowest $35\%$ of link speed values as congested and the other $65\%$ as free-flowing. In order to create $3$ classes, we use thresholds at $35\%$ as congested, $35\%$ to $60\%$ as mildly congested and the remaining $40\%$ as free-flowing. The temporal graph classification datasets can be found with our code.

%\BC{Soham, add your details here for how to convert traffic networks to temporal graph classification setting.}

% DynHCP datasets were derived by sliding a 30 s (50‐timepoint) window with a stride of 3 over the 150‐point ROI time series for 200 cortical regions.  Within each window, edges are weighted by the Pearson correlation between regional time series, yielding $\frac{150 - 50}{3} + 1 =34$ snapshots per subject.

\subsection{Temporal Degree Encoding for Node Features} \label{app:temporal-degree}
In datasets that lack predefined node features, such as social network datasets, we introduce a novel node representation called \textit{temporal degree}, which captures the temporal evolution of node connectivity. This representation is used as the initial input embedding for the GraphSAGE component in our model. The temporal degree assigns a vector to each node, where each entry corresponds to the node’s degree at a specific time step. For example, in \Cref{fig:temporal}, the temporal degree vector for node $A$ is $[2,1,1,0,0,0]$, indicating that it has two edges at time step 1, one edge at time step 2, and one edge at time step 3. Similarly, the vector for node $B$ is $[1,0,1,1,0,1]$. 

For datasets such as brain networks and traffic networks that come with intrinsic node features, we leverage those domain-specific attributes directly. However, in the absence of such features, our temporal degree encoding provides a meaningful and informative alternative that enhances the model’s ability to learn from temporal graph dynamics.

\subsection{Runtime}
Our model, \textsc{T3Former}, achieves high accuracy across temporal graph classification tasks while maintaining low computational complexity. The detailed runtime breakdown for each step, along with the total runtime, is provided in \Cref{tab:runtime}. Feature extraction was conducted on a machine with an Intel i7 13th Gen processor and 16 GB RAM, while the training and evaluation of \textsc{T3Former} were performed on the system described in the training details section. This analysis highlights the efficiency of our approach in both feature computation and model training.

\subsection{t-SNE Visualizations} \label{app:tsne}

\begin{figure}[t]
    \centering
    % First row
    \begin{subfigure}[b]{0.47\linewidth}
        \centering
        \includegraphics[width=\linewidth]{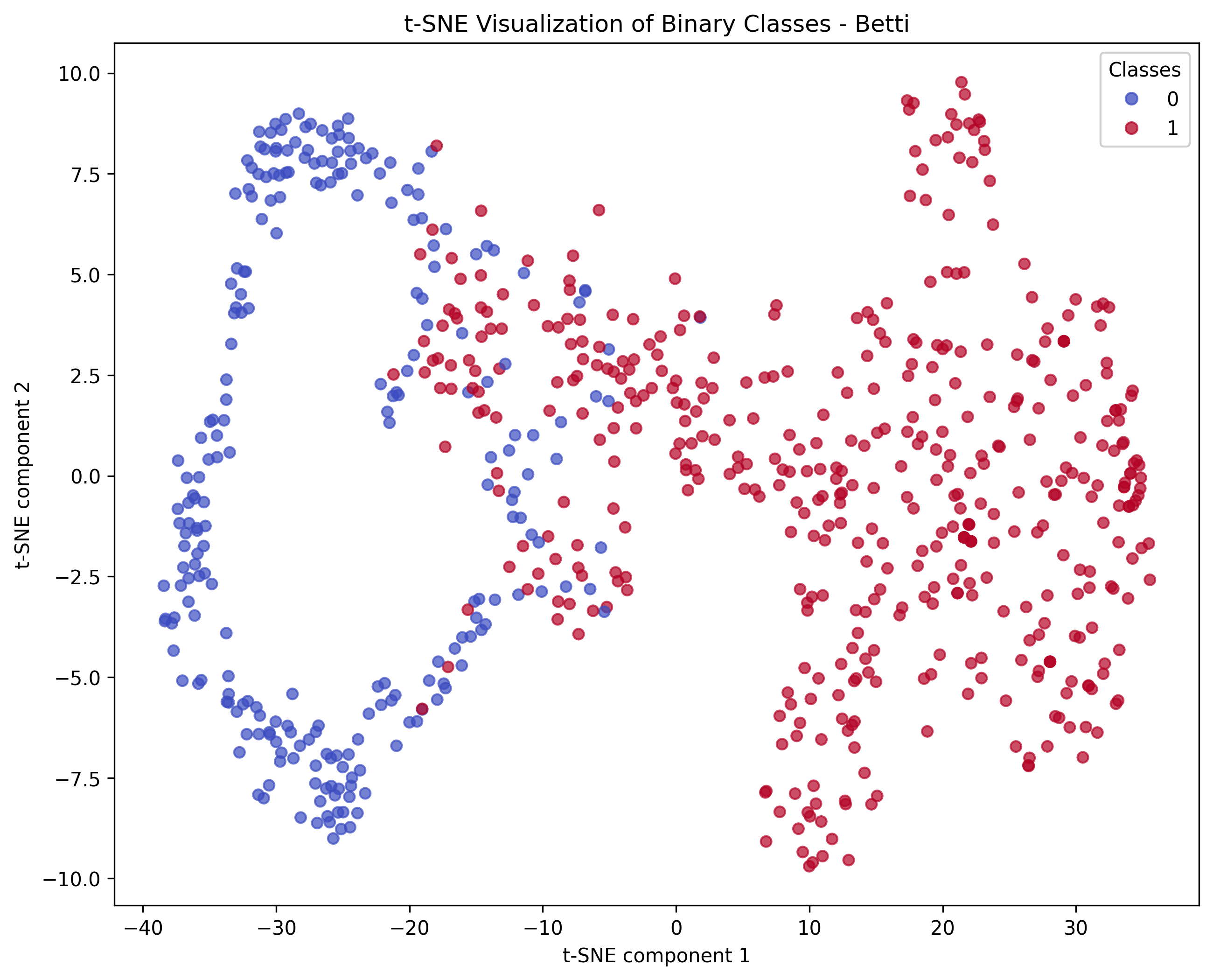}
        \caption{Betti vectors}
    \end{subfigure}
 %   \hspace{0.04\linewidth}
    \begin{subfigure}[b]{0.47\linewidth}
        \centering
        \includegraphics[width=\linewidth]{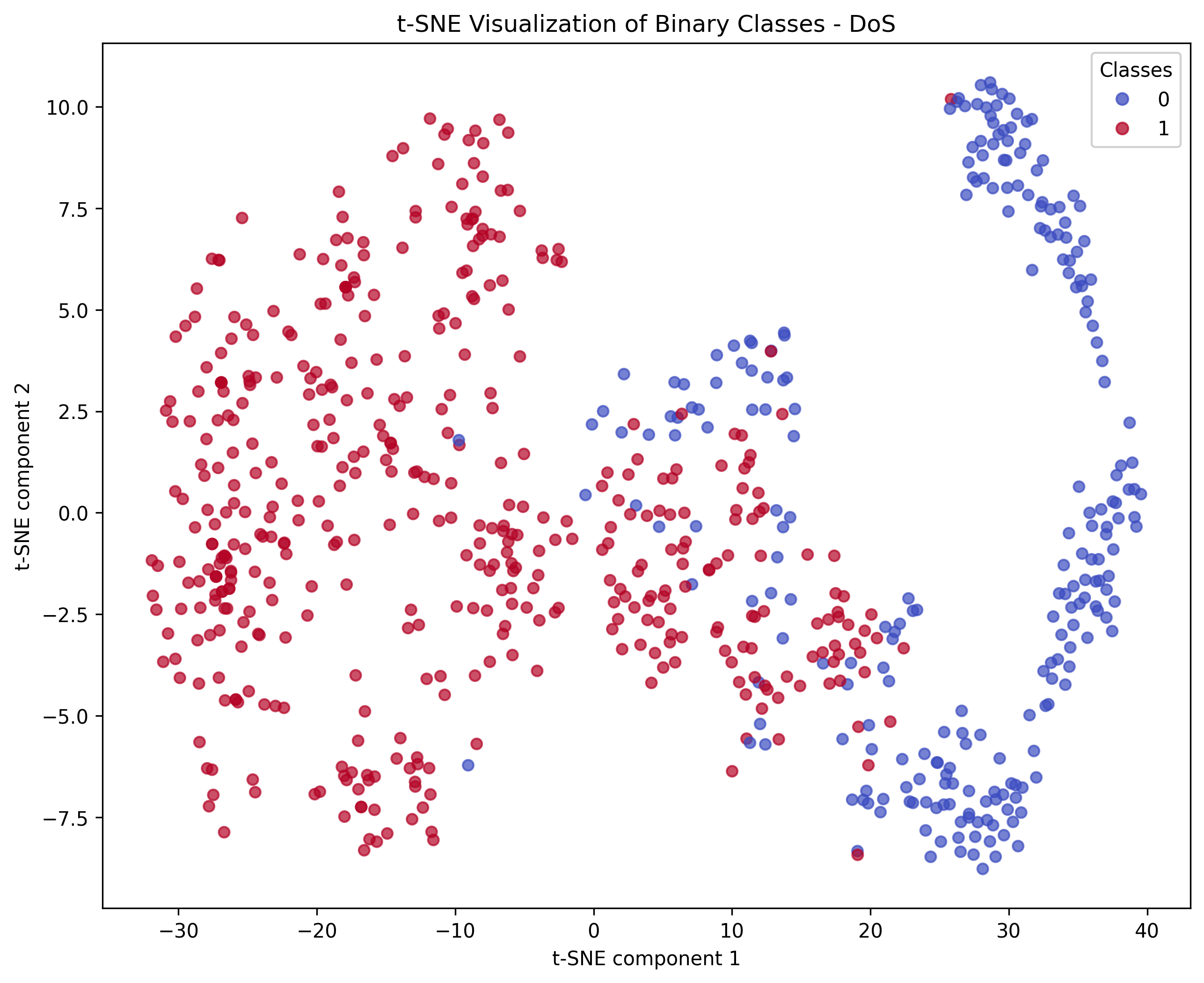}
        \caption{DoS vectors}
    \end{subfigure}
    
    \begin{subfigure}[b]{0.47\linewidth}
        \centering
        \includegraphics[width=\linewidth]{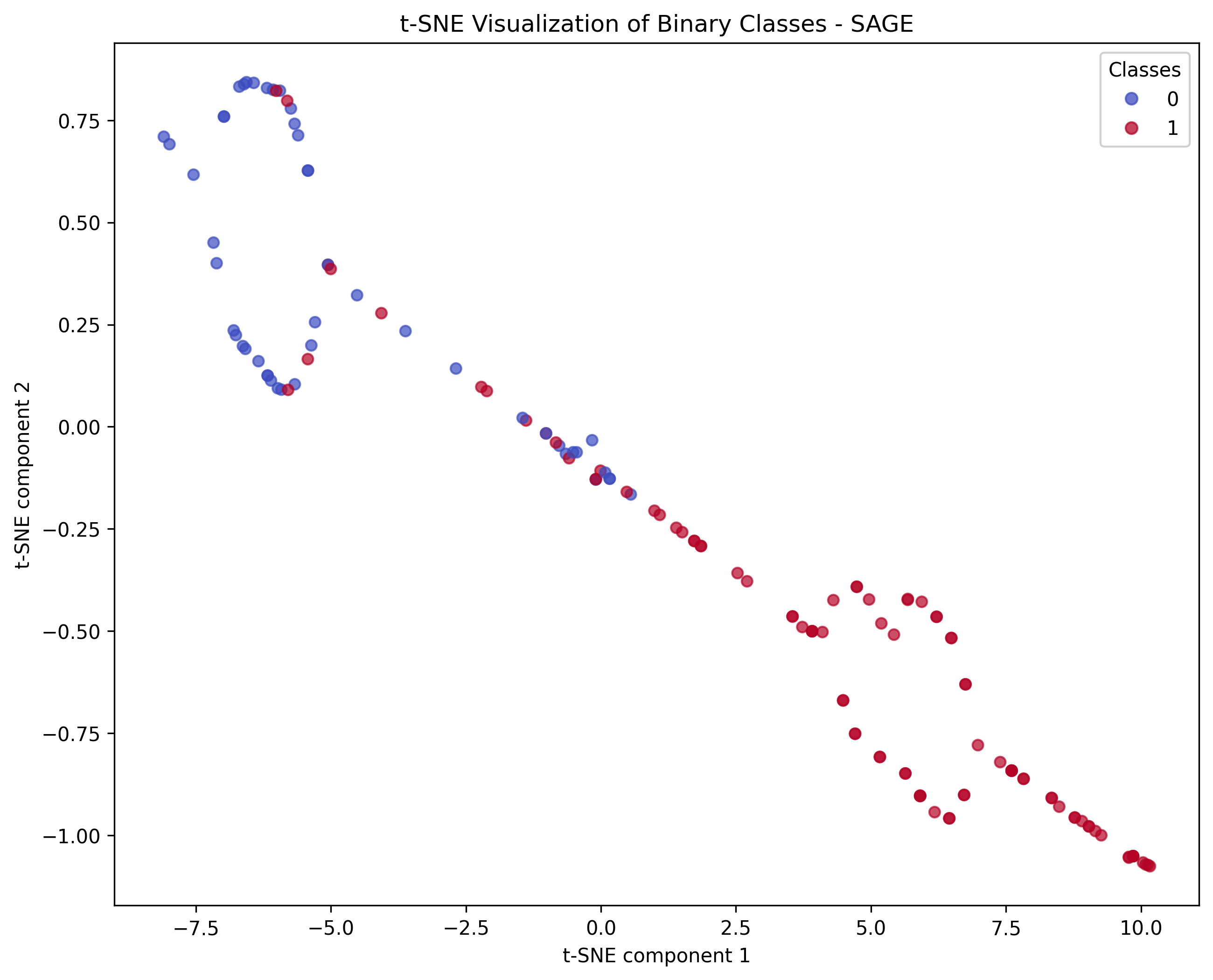}
        \caption{SAGE embeddings}
    \end{subfigure}
    %\hspace{0.04\linewidth}
    \begin{subfigure}[b]{0.47\linewidth}
        \centering
        \includegraphics[width=\linewidth]{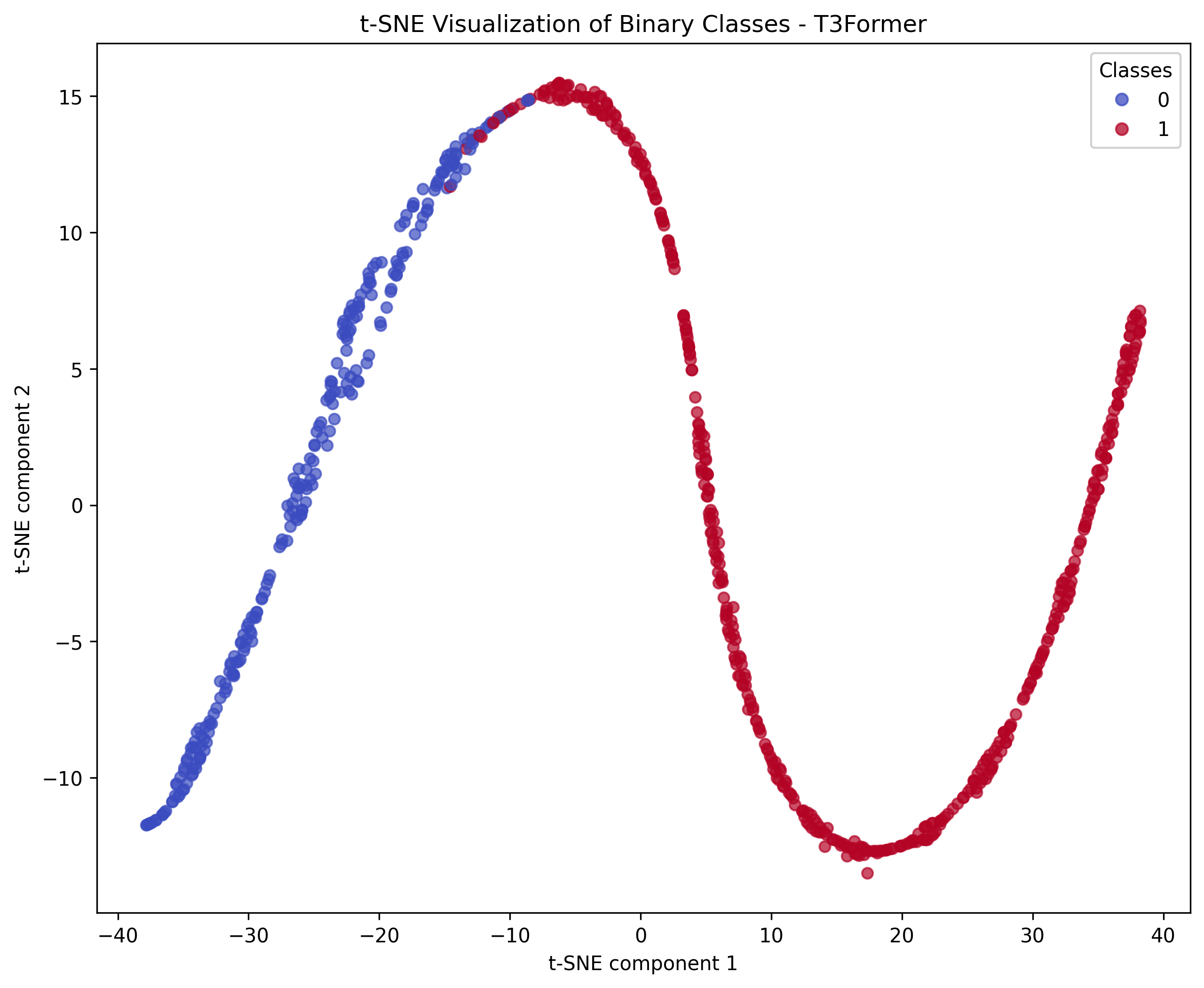}
        \caption{T3former}
    \end{subfigure}
    \caption{\footnotesize {\bf PEMS04 (2) tSNE Visualizations.} t-SNE plots of individual embedding representations from T3former components on the PEMS04 traffic dataset with two classes.}
    \label{fig:tsne-pems2}
   \vspace{-.1in}
\end{figure}

\begin{figure}[h!]
   %\vspace{-.1in}
    \centering
    % First row
    \begin{subfigure}[b]{0.47\linewidth}
        \centering
        \includegraphics[width=\linewidth]{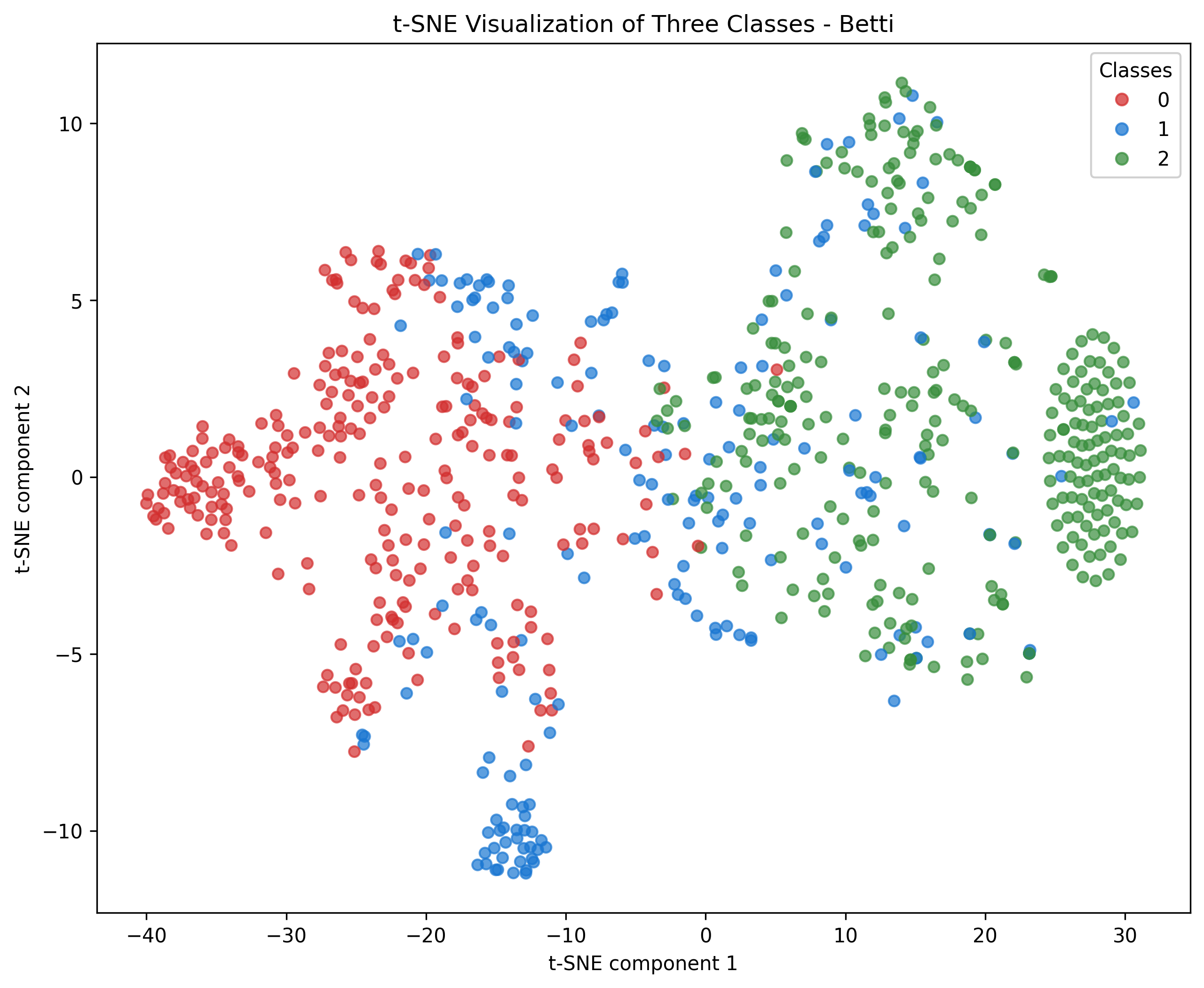}
        \caption{Betti vectors}
    \end{subfigure}
 %   \hspace{0.04\linewidth}
    \begin{subfigure}[b]{0.47\linewidth}
        \centering
        \includegraphics[width=\linewidth]{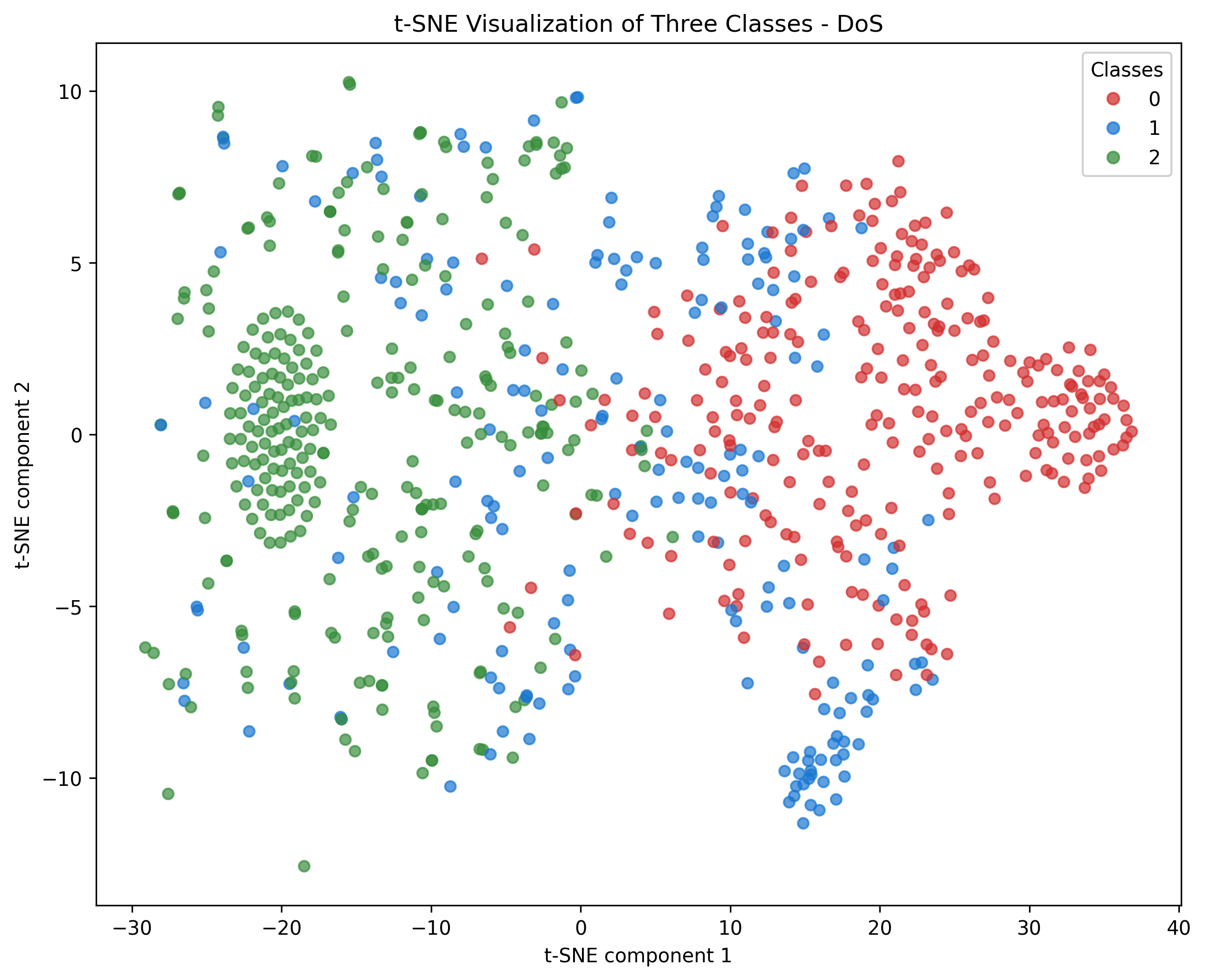}
        \caption{DoS vectors}
    \end{subfigure}
    
    \begin{subfigure}[b]{0.47\linewidth}
        \centering
        \includegraphics[width=\linewidth]{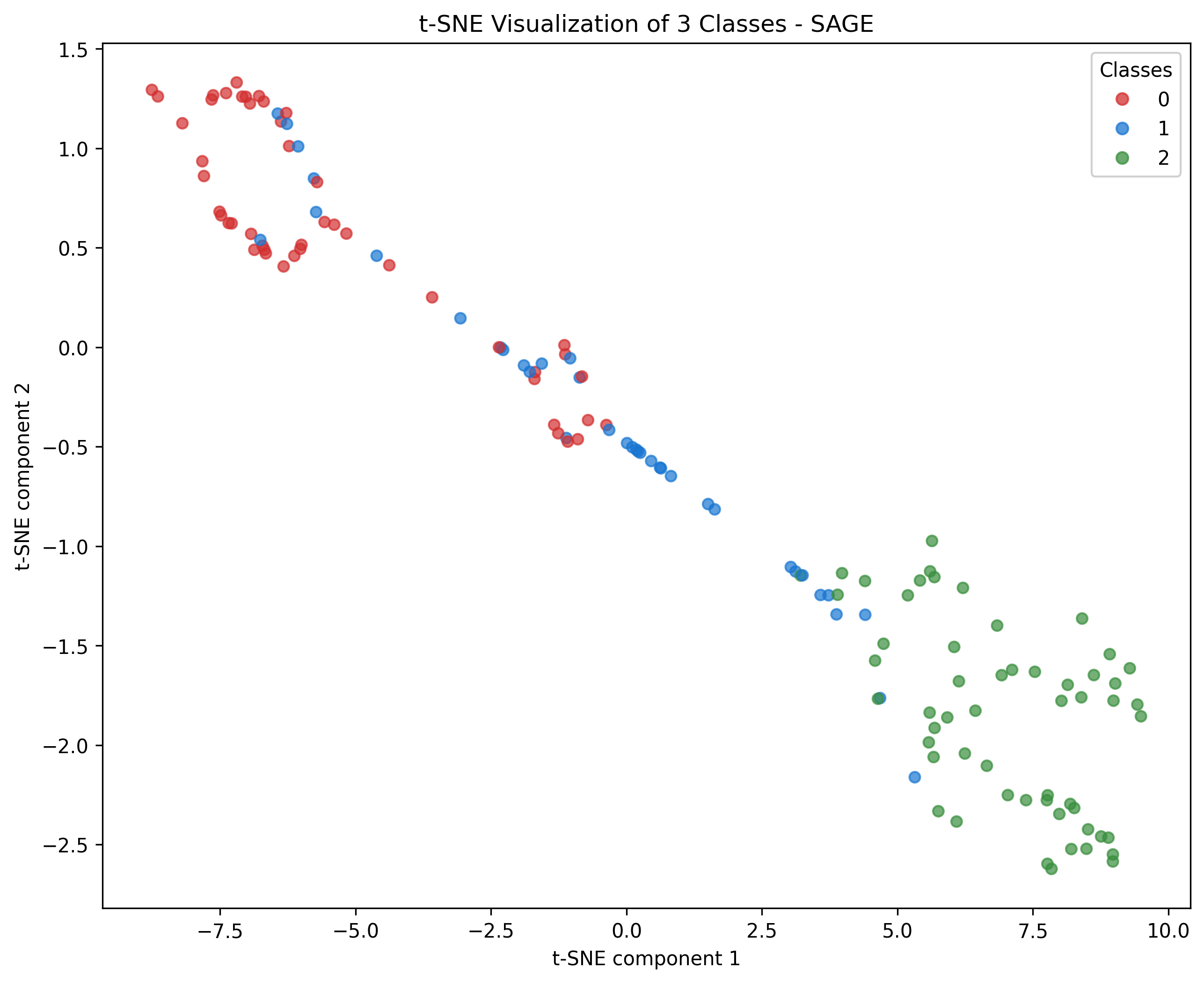}
        \caption{SAGE embeddings}
    \end{subfigure}
    %\hspace{0.04\linewidth}
    \begin{subfigure}[b]{0.47\linewidth}
        \centering
        \includegraphics[width=\linewidth]{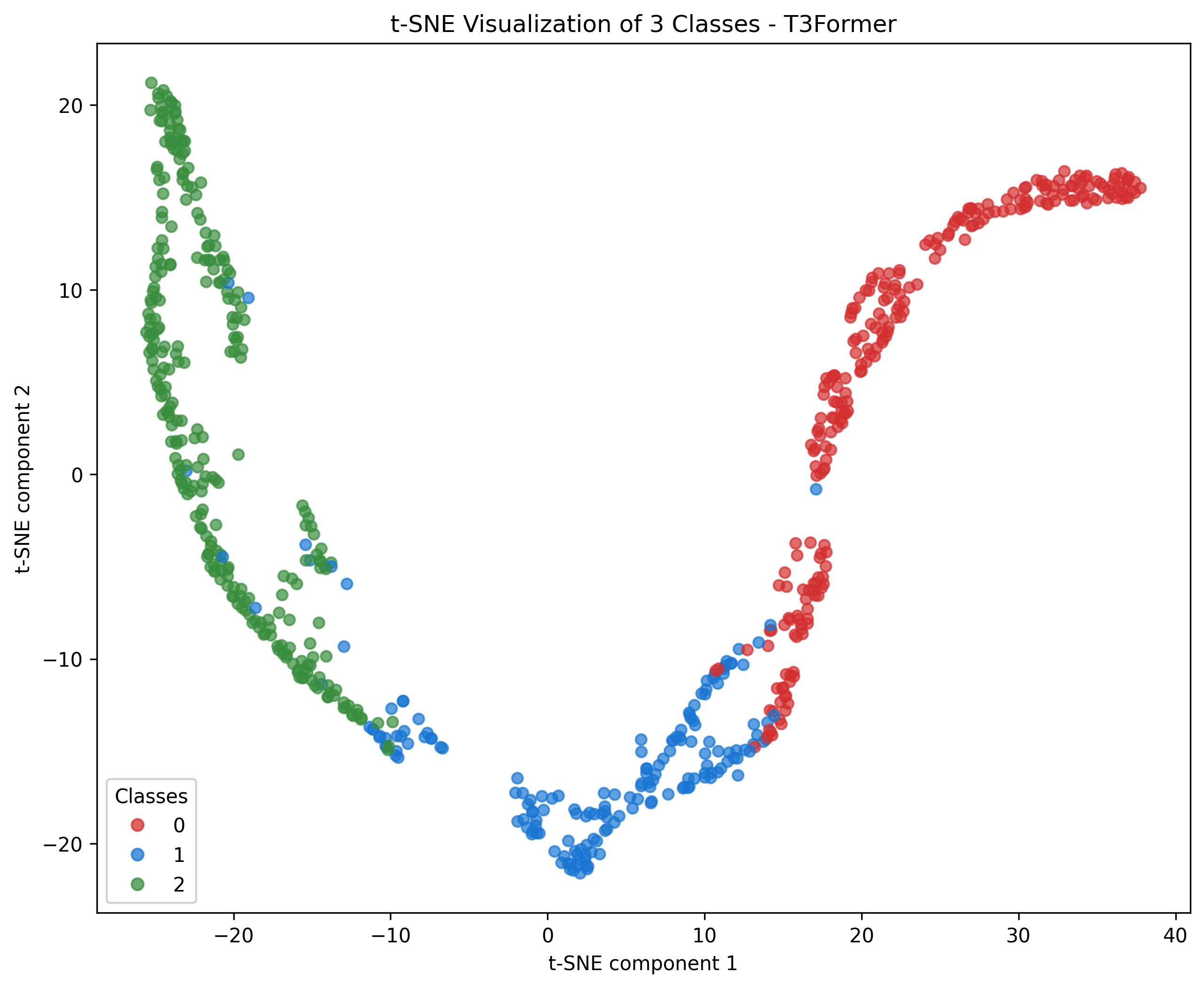}
        \caption{T3former}
    \end{subfigure}

    \caption{\footnotesize {\bf PEMS04 (3) tSNE Visualizations.} t-SNE plots of individual embedding representations from T3former components on the PEMS04 traffic dataset with three classes.}
    \label{fig:tsne-pems3}
\end{figure}

\begin{figure}[h!]
    \centering
    % First row
    \begin{subfigure}[b]{0.47\linewidth}
        \centering
        \includegraphics[width=\linewidth]{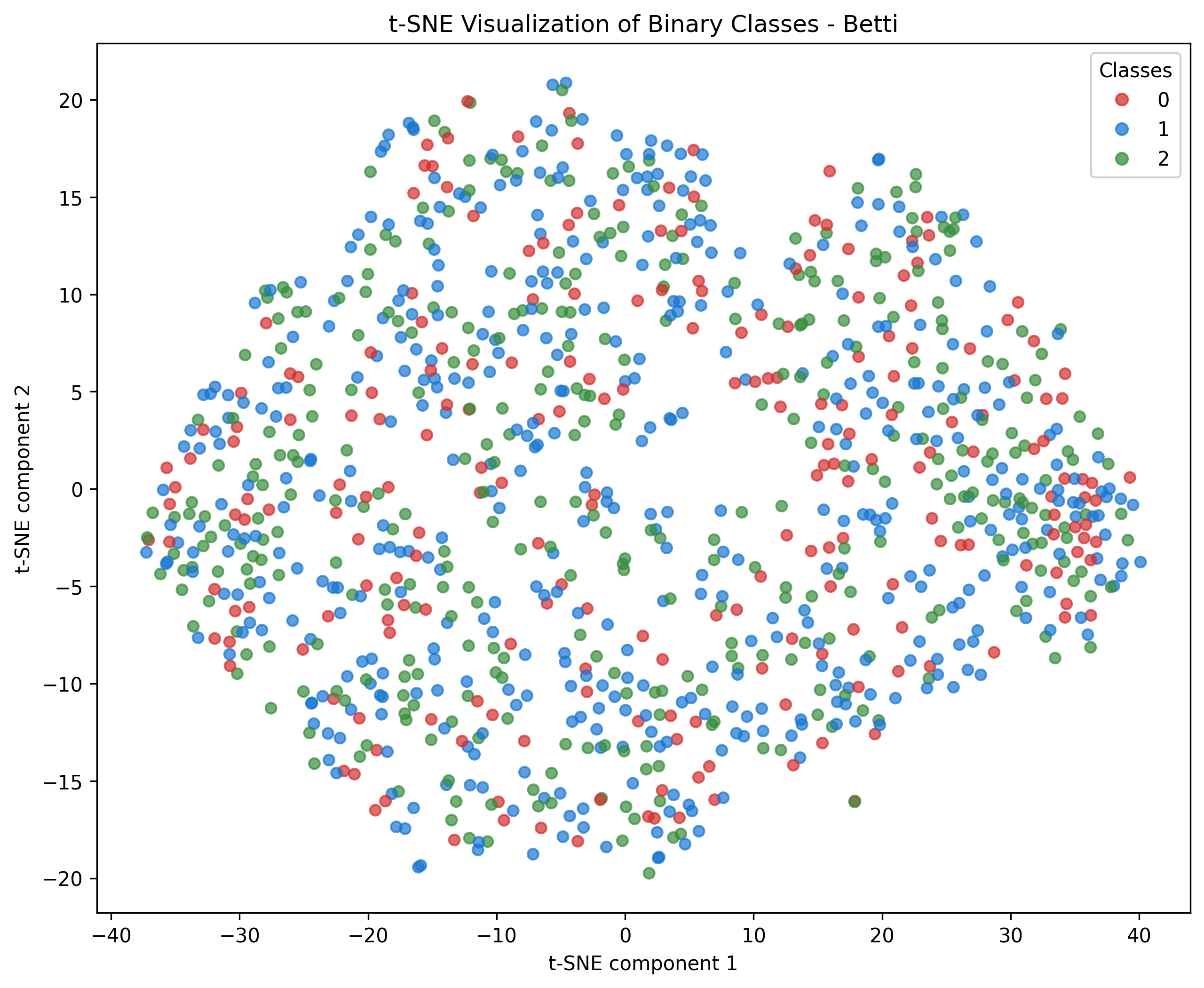}
        \caption{Betti vectors}
    \end{subfigure}
 %   \hspace{0.04\linewidth}
    \begin{subfigure}[b]{0.47\linewidth}
        \centering
        \includegraphics[width=\linewidth]{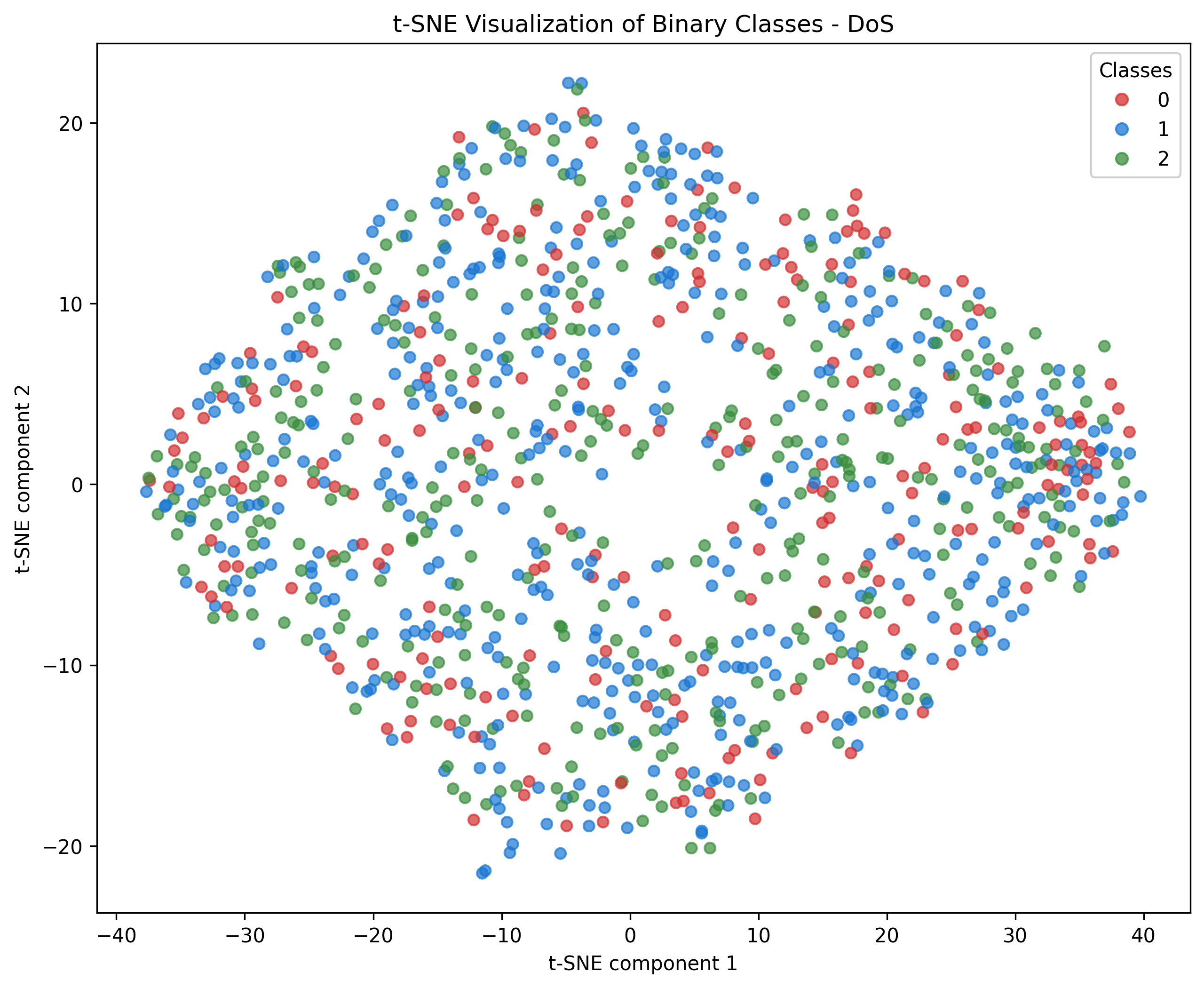}
        \caption{DoS vectors}
    \end{subfigure}
    
    \begin{subfigure}[b]{0.47\linewidth}
        \centering
        \includegraphics[width=\linewidth]{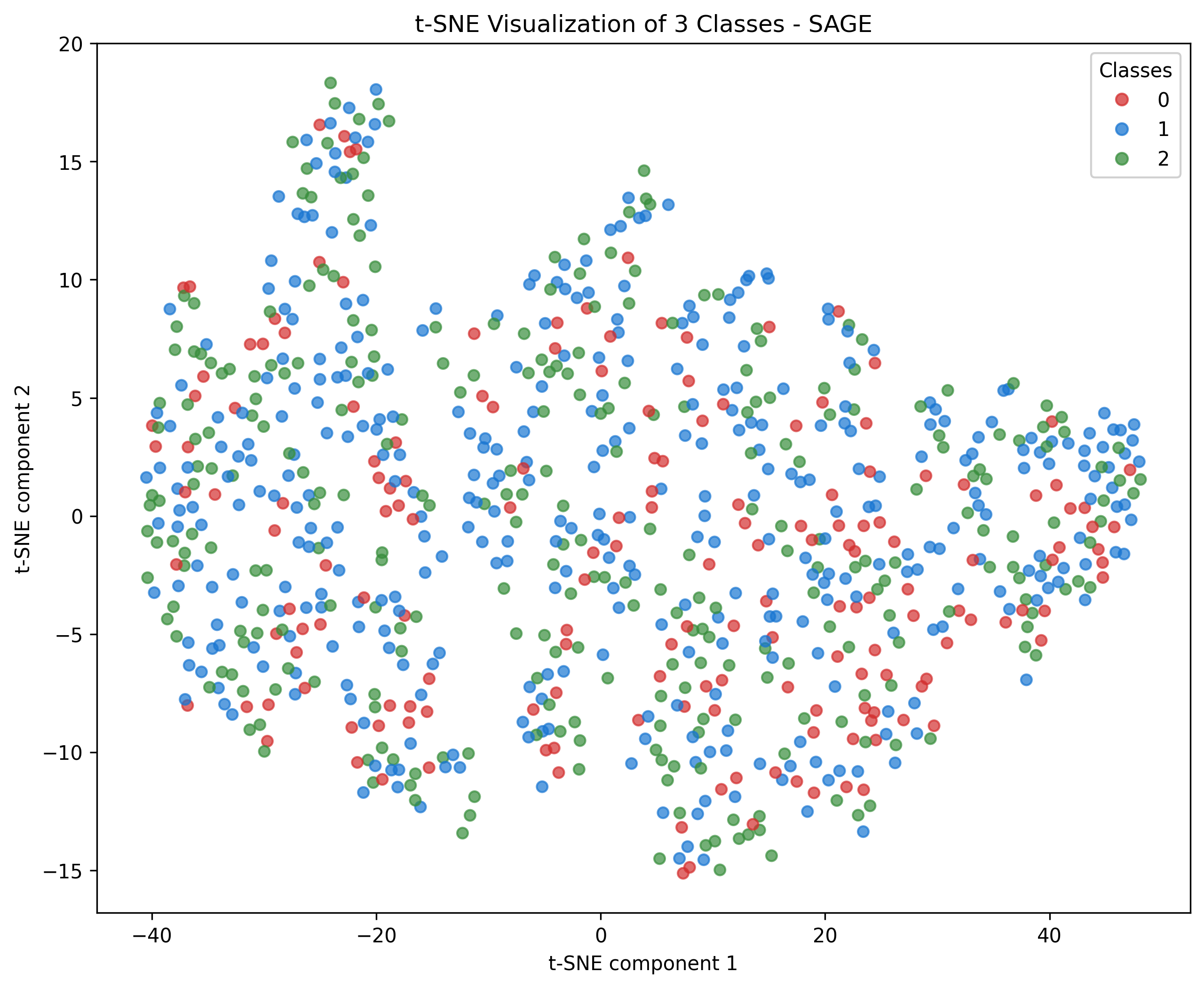}
        \caption{SAGE embeddings}
    \end{subfigure}
    %\hspace{0.04\linewidth}
    \begin{subfigure}[b]{0.47\linewidth}
        \centering
        \includegraphics[width=\linewidth]{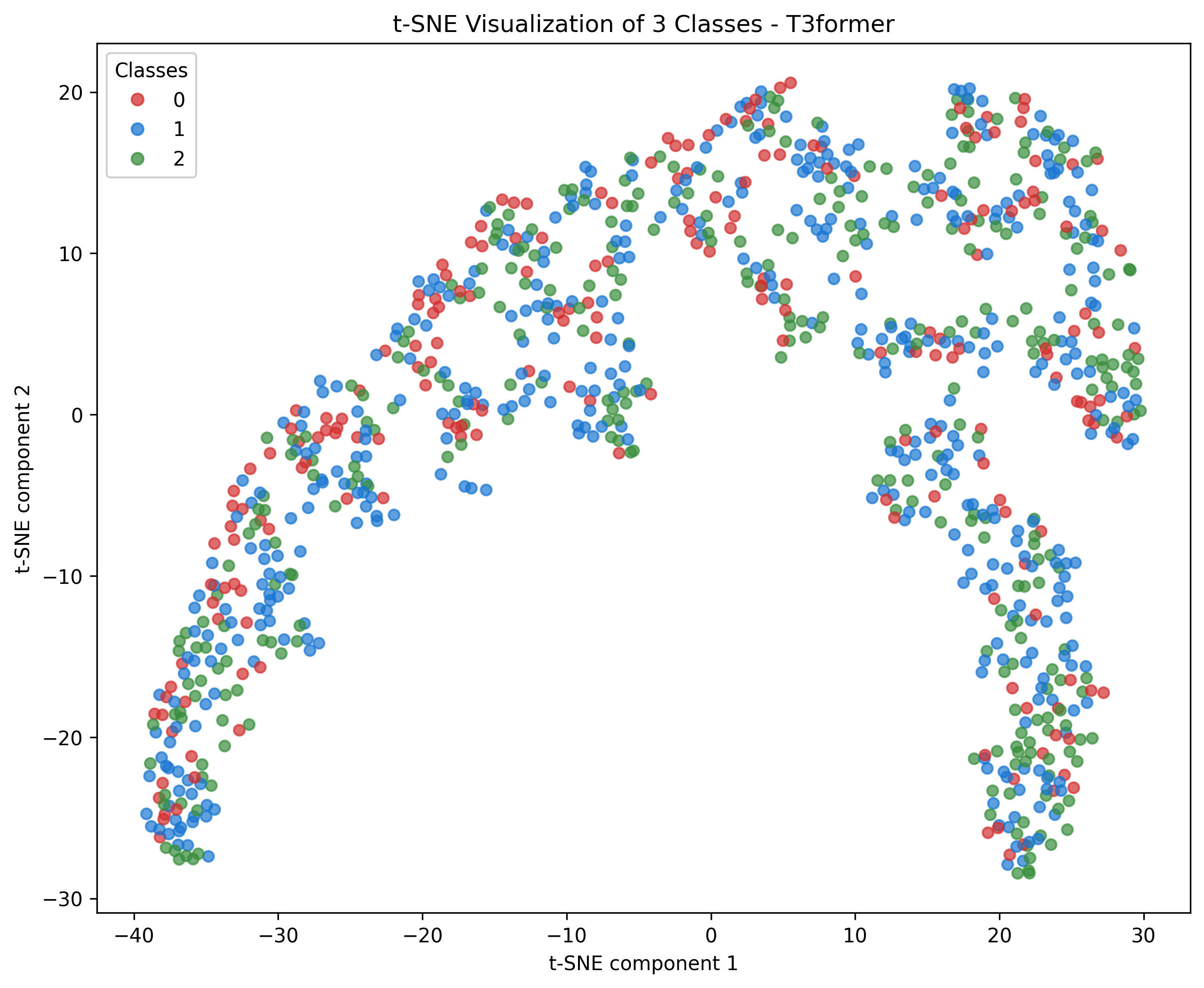}
        \caption{T3former}
    \end{subfigure}

    \caption{{\bf DynHCP-Age tSNE Visualizations.} t-SNE plots of individual embedding representations from T3former components on the DynAge brain connectivity dataset with three classes.}
    \label{fig:tsne-age}
\end{figure}

The t-SNE visualizations (\Cref{fig:tsne-pems2,fig:tsne-pems3,fig:tsne-age}) for PEMS04 (2-class and 3-class) and DynAge datasets illustrate the discriminative power of the embeddings extracted by different T3former components. In both PEMS04 scenarios, Betti and DoS vectors reveal clear class separations, highlighting their effectiveness in capturing underlying structural and spectral differences. In contrast, time-agnostic SAGE embeddings show limited separation, indicating less sensitivity to class-specific structural patterns. 

%We created t-Stochastic Neighbor Embedding (t-SNE) plots for the different embeddings in our model \textsc{T3former}, along with the final embeddings. This illustrates how what we see in Table $5$, as T3former creates embeddings that classify temporal graphs with a higher accuracy.

 The unified T3former embeddings demonstrate the clearest and most structured class distinctions, forming smooth nonlinear manifolds, which implies successful integration of temporal, structural, and spectral features. For the DynAge dataset, classes are inherently more intertwined due to higher complexity, yet T3former embeddings still provide better separability and structure compared to individual descriptors or baseline embeddings, confirming the advantage of a hybrid approach.

\subsection{Sliding Windows} \label{app:sliding-window}

\paragraph{Sliding Window Construction.} In this part, we provide a detailed description and toy example of our sliding-window construction for temporal graph snapshots.
Formally, let $\mathcal{G} = (\mathcal{V}, \mathcal{E}, \tau)$
denote a temporal graph, where  $\tau:\mathcal{E}\to\mathbb{R}^+$ assigns to each edge the time of its occurrence (or a multiset of times if it appears repeatedly).  

Given a window length $\delta>0$, we extract a family of induced subgraphs 
\[
G_{[t,\,t+\delta]} = \bigl(\mathcal{V}_t,\;\mathcal{E}_t\bigr)
\]
parameterized by the window start $t$, where
\smallskip

\centerline{$\mathcal{E}_t 
  = \{\,e\in\mathcal{E}\mid \tau(e)\in[t,\,t+\delta]\},$}
 \smallskip
 
  \centerline{$\mathcal{V}_t 
  = \{\,u\in\mathcal{V}\mid \exists\,e\in\mathcal{E}_t:\,u\in e\}.$}
  \smallskip

To obtain a sequence of such snapshots, we introduce a stride $\sigma\in(0,\delta)$, and define
\[
\G_i \;=\; G_{[\,i\,\sigma,\;i\,\sigma+\delta\,]}, 
\quad i = 0,1,2,\dots
\]
so that consecutive windows overlap by $\delta - \sigma$. In \Cref{fig:temporal}, we give a toy example of this construction for $\delta=2$ and $\sigma=1$.

\begin{figure*}[t]
    \centering
    \begin{subfigure}[b]{0.4\textwidth}
        \centering
        \includegraphics[width=\textwidth]{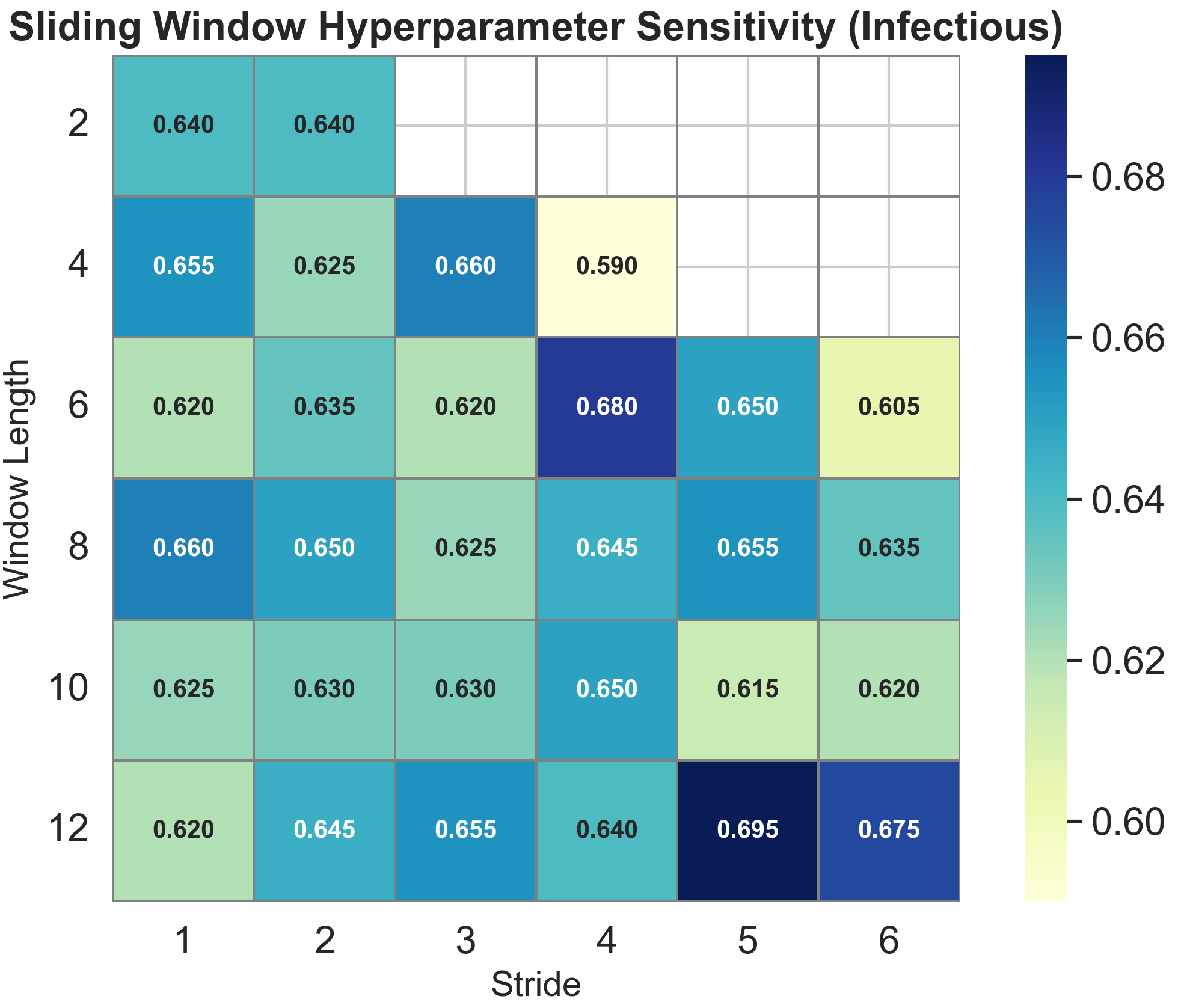}
        \caption{Infectious - Betti}
        \label{fig:infectious-Betti}
    \end{subfigure}
    \qquad
    \begin{subfigure}[b]{0.4\textwidth}
        \centering
        \includegraphics[width=\textwidth]{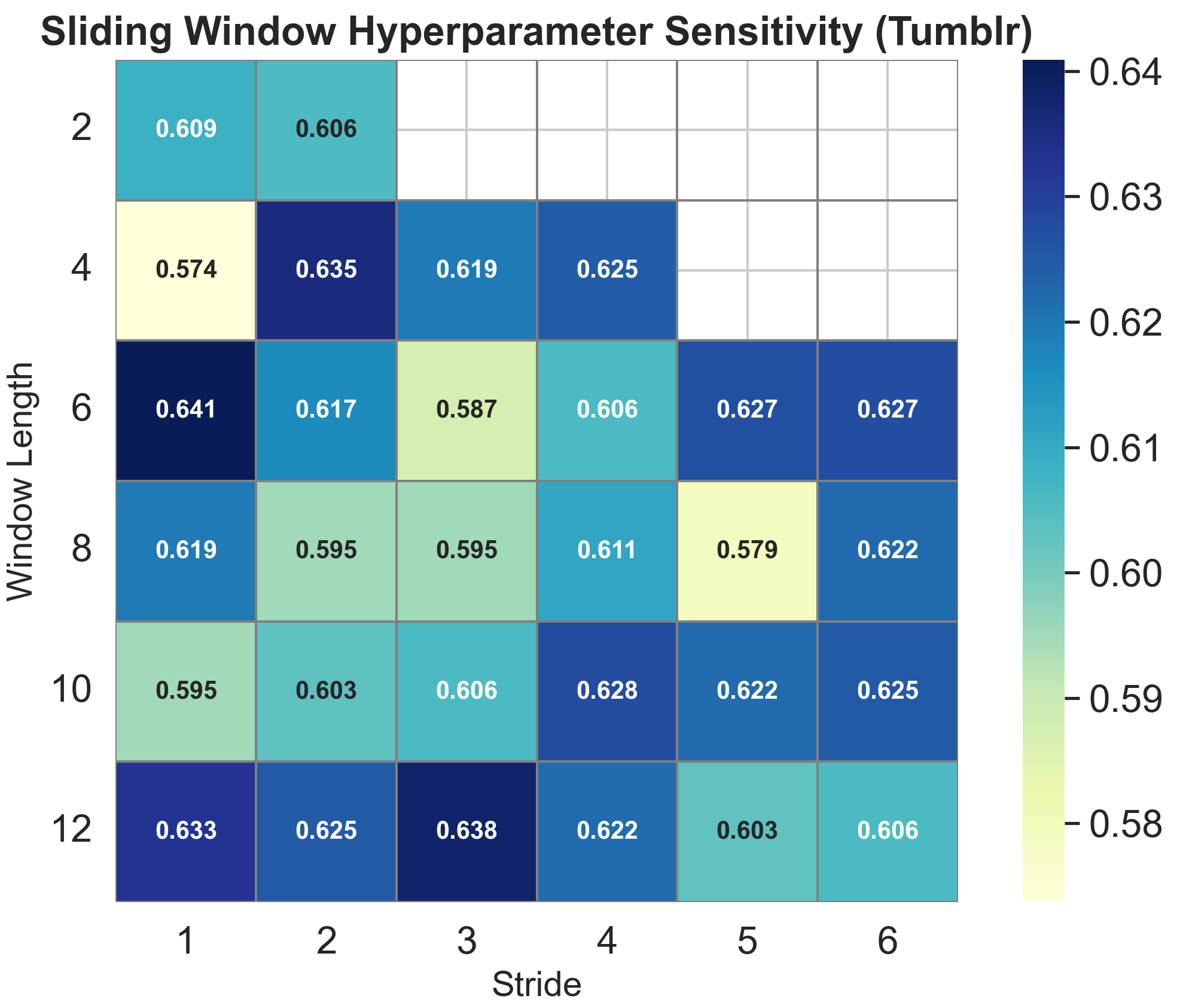}
        \caption{Tumblr - Betti}
        \label{fig:tumblr-Betti}
    \end{subfigure}

\begin{subfigure}[b]{0.4\textwidth}
        \centering
        \includegraphics[width=\textwidth]{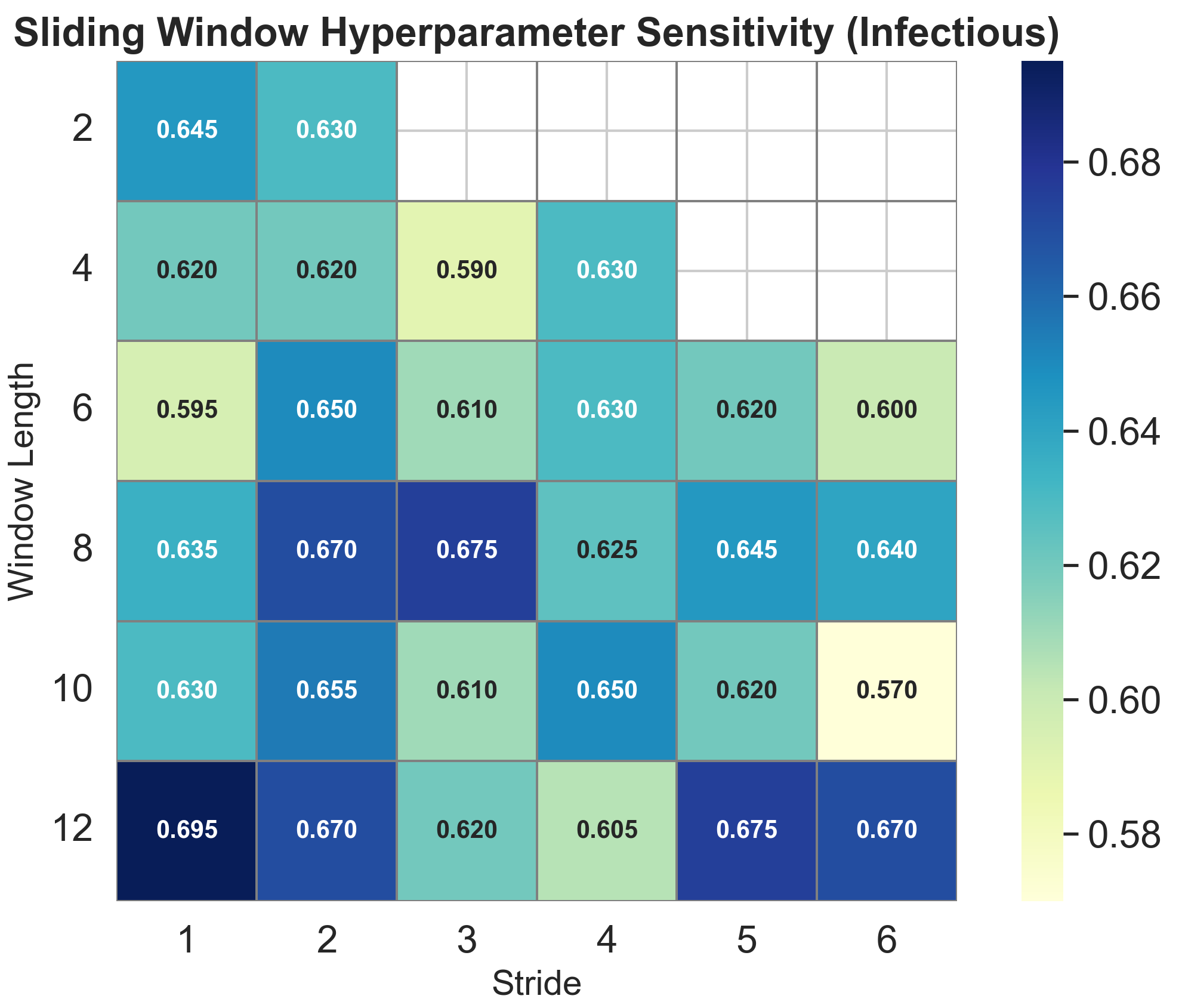}
        \caption{Infectious - DoS}
        \label{fig:infectious}
    \end{subfigure}
\qquad    \begin{subfigure}[b]{0.4\textwidth}
        \centering
        \includegraphics[width=\textwidth]{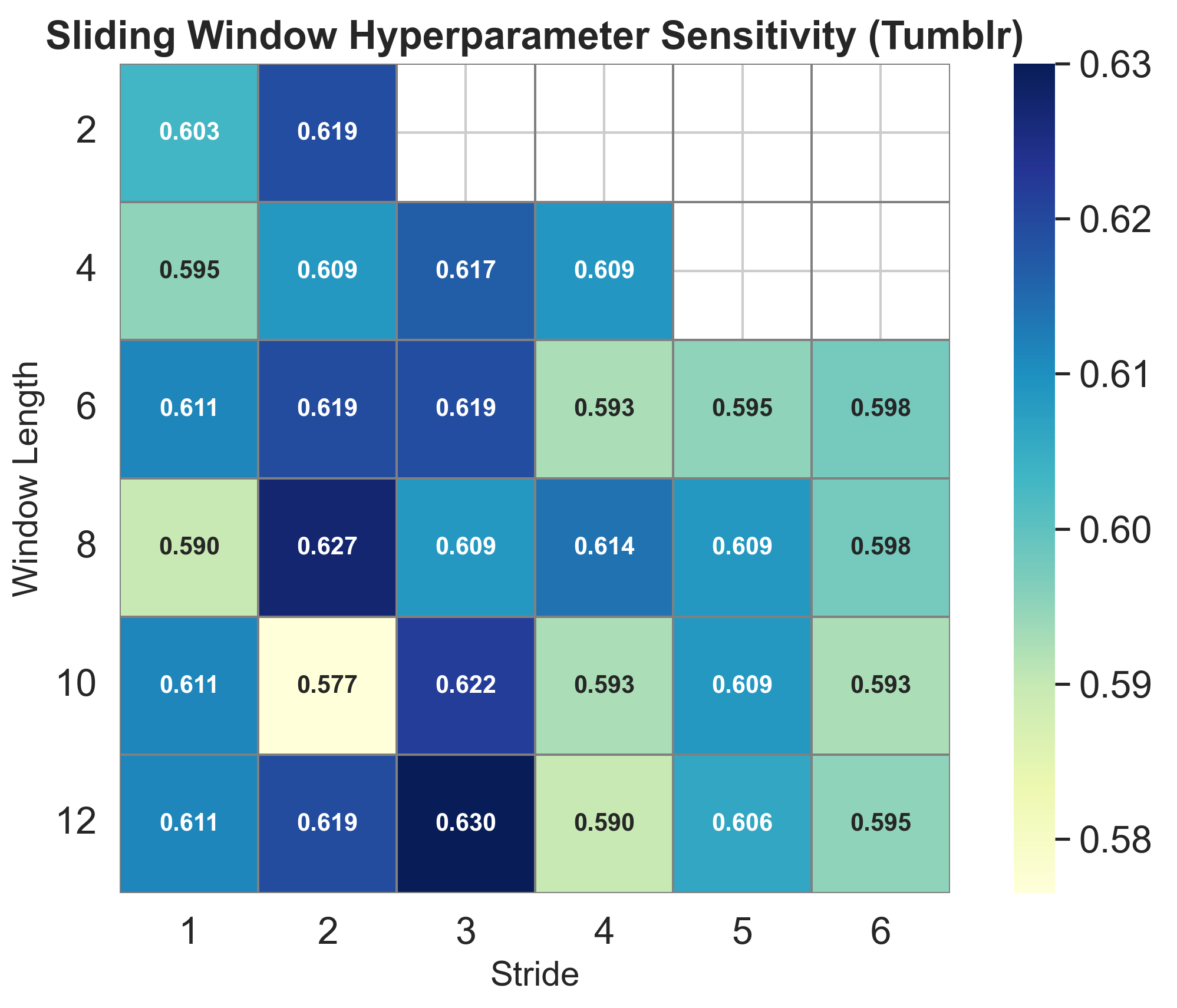}
        \caption{Tumblr - DoS}
        \label{fig:tumblr}
    \end{subfigure}
    
\caption{\textbf{Sliding-Window Hyperparameter Sensitivity.}  
Heatmaps show how varying the window length (y–axis) and stride (x–axis) affects classification accuracy of Betti and DoS vectors on two temporal social network datasets: \texttt{Infectious} and \texttt{Tumblr}. Each cell reports the mean accuracy for the corresponding parameter setting.}
    \label{fig:sliding-window-sensitivity2}
\end{figure*}

\paragraph{Hyperparameter Analysis.} To select the optimal sliding‐window hyperparameters, we performed a grid search over window length and stride. \Cref{fig:sliding-window-sensitivity2} report the resulting 5-fold cross‐validated accuracy for Density of States features and Betti vectors for \texttt{INFECTIOUS} and \texttt{TUMBLR} datasets. As these plots demonstrate stable performance across settings, we fix a window length of 6 and a stride of 4 for all experiments.

